\documentclass{article}

\usepackage{microtype}
\usepackage{graphicx}
\usepackage{subcaption}
\usepackage{booktabs} 
\usepackage{multirow} 
\usepackage{wrapfig} 
\usepackage{enumitem} 

\usepackage{hyperref}

\usepackage[preprint]{icml2026}

\usepackage{amsmath}
\usepackage{amssymb}
\usepackage{mathtools}
\usepackage{amsthm}

\renewcommand{\vec}[1]{\mathbf{#1}}

\usepackage[capitalize,noabbrev]{cleveref}

\usepackage{listings}
\usepackage{xcolor}

\definecolor{codegreen}{RGB}{40, 120, 40}
\definecolor{codegray}{RGB}{100, 100, 100}
\definecolor{codeblue}{RGB}{0, 80, 160}
\definecolor{codebg}{RGB}{248, 248, 248}
\definecolor{codeframe}{RGB}{200, 200, 200}

\theoremstyle{plain}
\newtheorem{theorem}{Theorem}
\newtheorem{proposition}{Proposition}

\theoremstyle{definition}

\theoremstyle{remark}

\newcommand{\cD}{\mathcal{D}}
\newcommand{\E}{\mathbb{E}}
\DeclareMathOperator*{\argmin}{arg\,min}
\DeclareMathOperator{\Var}{Var}
\DeclareMathOperator{\var}{Var}
\DeclareMathOperator{\Cov}{Cov}
\DeclareMathOperator{\tr}{tr}
\newcommand{\Bp}[1]{(#1)}
\newcommand{\Sp}[1]{(#1)}
\newcommand{\Mp}[1]{[#1]}

\usepackage[textsize=tiny]{todonotes}

\newlength{\secvspace}
\let\originalsection\section
\let\originalsubsection\subsection
\let\originalappendix\appendix

\makeatletter
\renewcommand{\section}{\@ifstar{\@sectionstar}{\@sectionnormal}}
\newcommand{\@sectionnormal}[1]{%
  \vspace{-\secvspace}%
  \originalsection{#1}%
  \vspace{-\secvspace}%
}
\newcommand{\@sectionstar}[1]{%
  \vspace{-\secvspace}%
  \originalsection*{#1}%
  \vspace{-\secvspace}%
}

\renewcommand{\subsection}{\@ifstar{\@subsectionstar}{\@subsectionnormal}}
\newcommand{\@subsectionnormal}[1]{%
  \vspace{-\secvspace}%
  \originalsubsection{#1}%
  \vspace{-\secvspace}%
}
\newcommand{\@subsectionstar}[1]{%
  \vspace{-\secvspace}%
  \originalsubsection*{#1}%
  \vspace{-\secvspace}%
}

\newcommand{\restorenormalspacing}{%
  \setlength{\secvspace}{0pt}%
}

\renewcommand{\appendix}{%
  \restorenormalspacing%
  \originalappendix%
}
\makeatother

\usepackage[most]{tcolorbox}
\definecolor{myGreen}{RGB}{230,245,230}
\definecolor{myDarkGreen}{RGB}{60,120,60}
\definecolor{darkgreen}{RGB}{0,100,0}

\definecolor{propBg}{RGB}{235,245,255}
\definecolor{propFrame}{RGB}{70,130,180}
\newtcolorbox{propositionbox}{
    colback=propBg,
    colframe=propFrame,
    boxrule=0.8pt,
    arc=2mm,
    left=3mm,
    right=3mm,
    top=2mm,
    bottom=2mm
}

\definecolor{thmBg}{RGB}{235,250,240}
\definecolor{thmFrame}{RGB}{60,140,90}
\newtcolorbox{theorembox}{
    colback=thmBg,
    colframe=thmFrame,
    boxrule=0.8pt,
    arc=2mm,
    left=3mm,
    right=3mm,
    top=2mm,
    bottom=2mm
}

\newtcolorbox{simplebox}{
    colback=gray!5,
    colframe=gray!50,
    boxrule=0.5pt,
    arc=2mm,
    left=3mm,
    right=3mm,
    top=2mm,
    bottom=2mm
}

\newtcolorbox{responsebox}[1][]{
    colback=gray!5,
    colframe=gray!60,
    boxrule=0.5pt,
    arc=2mm,
    left=3mm,
    right=3mm,
    top=2mm,
    bottom=2mm,
    title={#1},
    fonttitle=\bfseries
}

\icmltitlerunning{Shrinking the Variance: Shrinkage Baselines for Reinforcement Learning with Verifiable Rewards}

\begin{document}

\twocolumn[
  \icmltitle{Shrinking the Variance: Shrinkage Baselines for Reinforcement Learning with Verifiable Rewards}



  \icmlsetsymbol{equal}{*}
  \icmlsetsymbol{theory}{\dag}

  \begin{icmlauthorlist}
    \icmlauthor{Guanning Zeng}{cmu}
    \icmlauthor{Zhaoyi Zhou}{cmu,theory}
    \icmlauthor{Daman Arora}{cmu}
    \icmlauthor{Andrea Zanette}{cmu}
  \end{icmlauthorlist}

  \icmlaffiliation{cmu}{Carnegie Mellon University}
  \icmlcorrespondingauthor{Andrea Zanette}{zanette@cmu.edu}
  \icmlcorrespondingauthor{Guanning Zeng}{zgn0303@gmail.com}

  \icmlkeywords{Reinforcement Learning, Maximum Likelihood, Large Language Models, Reasoning}

  \vskip 0.3in

]



\printAffiliationsAndNotice{\icmlTheoryContribution}

\begin{abstract}

  Reinforcement Learning with Verifiable Rewards (RLVR) has emerged as a powerful paradigm for post-training large reasoning models (LRMs) using policy-gradient methods such as GRPO. To stabilize training, these methods typically center trajectory rewards by subtracting the empirical mean for each prompt. Statistically, this centering acts as a \emph{control variate} (or \emph{baseline}), reducing the variance of the policy-gradient estimator.
  
  Typically, the mean reward is estimated using per-prompt empirical averages for each prompt in a batch. Drawing inspiration from Stein’s paradox, we propose using \emph{shrinkage estimators} that combine \emph{per-prompt} and \emph{across-prompt} means to improve the overall per-prompt mean estimation accuracy---particularly in the low-generation regime typical of RLVR. Theoretically, we construct a shrinkage-based baseline that provably yields lower-variance policy-gradient estimators across algorithms. Our proposed baseline serves as a drop-in replacement for existing per-prompt mean baselines, requiring no additional hyper-parameters or computation. Empirically, shrinkage baselines consistently outperform standard empirical-mean baselines, leading to lower-variance gradient updates and improved training stability.

\end{abstract}

\vspace{-12pt}
\section{Introduction}
  
  Large reasoning models such as \textit{OpenAI-o1} \citep{openai_learning_2024} and \textit{DeepSeek-R1} \citep{guo2025deepseek} have demonstrated impressive reasoning capabilities, underscoring the effectiveness of reinforcement learning (RL) techniques for model post-training. A particularly impactful paradigm for fine-tuning reasoning models is \textbf{Reinforcement Learning with Verifiable Rewards (RLVR)}, where models are optimized using sparse, rule-based scalar rewards explicitly indicating the correctness of the model's final answer. RLVR-style training has shown substantial promise for tasks that require explicit and verifiable logic, such as mathematical or logical reasoning.
  
  A common approach to RLVR is applying policy gradient methods such as REINFORCE \citep{williams1992simple} or more modern variants such as GRPO \citep{shao2024deepseekmath} is optimized to maximize expected rewards through stochastic gradient estimates. A well-known challenge in policy gradient methods is the high variance of these gradient estimators  \citep{SuttonBarto2018-AC}, which can hinder stable training. To mitigate this, RL methods introduce a baseline---known in statics as a control variate---which is a state-dependent shift in the rewards that reduces variance in the gradient without introducing bias \citep{sutton1998reinforcement,greensmith2004variance}. The canonical choice is the value function, defined as the expected return from the state under consideration (or the initial state in RLVR) under the current policy. Despite being a heuristic rather than the theoretically optimal baseline, it is widely adopted \citep{SuttonBarto2018-AC} because it is straightforward to approximate while providing substantial variance reduction.
  
  Typically, the value function itself is unknown and it must be estimated.
  There are two broad classes of value function estimators. \emph{Classical RL approaches} introduce an auxiliary neural network to approximate the value function \citep{barto1989learning, mnih2016asynchronous, haarnoja2018soft, schulman2015trust, schulman2017proximal}. When carefully tuned, this strategy can be effective, but it comes with significant practical challenges: increased hyperparameter sensitivity, added engineering complexity, and the cost of training and maintaining an additional network for variance reduction. In contrast, \emph{recent methods for reasoning models}—including GRPO \citep{shao2024deepseekmath}, RLOO \citep{ahmadian2024back}, ReMax \citep{li2023remax}, REINFORCE++ \citep{hu2025reinforce++}, DAPO \citep{yu2025dapo} and CISPO \citep{minimax2025minimaxm1scalingtesttimecompute}—forgo explicit value function approximation. Instead, they construct baselines directly from Monte Carlo returns, typically using per-prompt empirical averages of generated responses. This avoids the overhead of a separate network and yields unbiased (or nearly unbiased) estimates, making it attractive for large reasoning models. 
  
  However, compared to classical domains, RL training on LLMs typically operates with large batch sizes but \emph{fewer rollouts per prompt} due to the high cost of inference. Even in industrial-scale LLM training~\citep{guo2025deepseek,deepseekai2025deepseekv32pushingfrontieropen}, the number of rollouts per prompt is often limited to 16 or fewer. This presents a new challenge: accurately estimating the baseline itself becomes difficult when the number of samples is small. 
  
  In this work, we revisit this problem of estimating value functions in RLVR. Although the per-prompt sample mean is an unbiased estimator, it treats each prompt independently. By recognizing that value functions must be estimated \textbf{simultaneously} across all prompts in a batch, we can construct an estimator for the value function---and thus the gradient---with strictly lower mean squared error (MSE). In particular, we propose a new baseline estimator inspired by the classical James–Stein shrinkage principle \citep{james1961estimation, stein1956inadmissibility}. This estimator reduces variance by trading a small amount of bias in the baseline for improved overall efficiency. Crucially, despite using a biased baseline, the \textbf{resulting policy gradient estimator remains unbiased} and enjoys provable variance reduction under standard assumptions. 
  
  Our proposed baseline introduces \textbf{no additional hyperparameters}, making it a simple drop-in replacement for existing critic-free RL methods. It relies solely on frequentist principles, without requiring assumptions about task difficulty, training data distributions, or model architectures. Importantly, the shrinkage baseline can be computed with \textbf{negligible computational overhead}. Extensive experiments across diverse models, tasks, and rollout settings demonstrate that the shrinkage baseline estimator consistently outperforms other common baselines in variance reduction. Furthermore, we observe a significant decrease in policy gradient variance, aligning with our theoretical predictions.  

\section{Preliminaries}
  
\label{sec:preliminary}
  
Let $\pi_\theta$ denote the language model parameterized by $\theta$. Given a prompt $x$ from distribution $\cD$, the model outputs response $y \sim \pi_\theta(\cdot\mid x)$ and receives a deterministic, verifiable reward $r(x,y)$ (e.g., correctness in math tasks). The RL objective is to maximize the expected reward
\begin{align*}
J(\theta):=\E_{x\sim \cD, y\sim \pi_\theta(\cdot\mid x)}[r(x,y)].
\end{align*}
The vanilla REINFORCE algorithm \citep{williams1992simple} derives the policy gradient as 
\begin{align*}
\nabla_\theta J(\theta)=\E_{x\sim \cD, y\sim \pi_\theta(\cdot\mid x)}[r(x,y)\nabla_\theta \log \pi_\theta(y\mid x)],
\end{align*}
so that the gradient can be estimated with one online sample: 
\begin{align*}
g^{\mathrm{vanilla}}(x,y;\theta)=r(x,y)\nabla_\theta \log \pi_\theta(y\mid x). 
\end{align*}
A scalar, prompt-dependent baseline $b(x) \in \mathbb R$ can be added to further reduce the gradient variance while keeping the gradient unbiased \citep{sutton1998reinforcement}:
\begin{align*}
g^{\mathrm{baseline}}(x,y;\theta):=(r(x,y)-b(x))\nabla_\theta \log \pi_\theta(y\mid x).
\end{align*}
More samples further reduce variance. Thus, at each RL step, it is common to sample $n$ prompts $\textstyle\vec x = (x_1,\ldots,x_n)$ i.i.d.\ from $\cD$. For each prompt $x_i$, the model generates $m$ responses $\textstyle\vec y_i = (y_i^1,\ldots,y_i^m)$ from $\pi_\theta(\cdot\mid x_i)$ with rewards $\textstyle r_i^j := r(x_i,y_i^j)$. Let $\textstyle\vec Y = (\vec y_1,\ldots,\vec y_n)$. The policy gradient is estimated by
\begin{align}
\boxed{g(\vec x, \vec Y;\theta)\;:=\;\tfrac{1}{nm}\textstyle\sum_{i,j} (r_i^j-b_i^j)\nabla_\theta \log \pi_\theta(y_i^j|x_i).}
    \label{eq:grad}
\end{align}
Here $b_i^j$ is the baseline for sample $(x_i,y_i^j)$. Baselines are typically prompt-dependent here, but we allow per-reward baselines to accommodate leave-one-out estimators like RLOO \citep{ahmadian2024back}.\footnote{RLOO computes each baseline by averaging the other rewards for the same prompt except the one being estimated, ensuring independence between $b_i^j$ and $r_i^j$ for unbiasedness.} 
  
Another requirement for baselines is unbiasedness. As shown in \Cref{prop:unbiased}, \Cref{eq:grad} is an unbiased estimate of the policy gradient $\nabla_\theta J(\theta)$, as long as $b_i^j$ and $r_i^j$ are independent for all $1\leq i\leq n, 1\leq j\leq m$.
\begin{propositionbox}
\begin{proposition}[Unbiasedness]
\label{prop:unbiased}
Suppose $b_i^j$ is independent of $y_i^j$ for all $i,j$. Then $g(\vec x, \vec Y;\theta)$ is unbiased, i.e., $\E[g(\vec x, \vec Y;\theta)] = \nabla_\theta J(\theta)$.
\end{proposition}
\end{propositionbox}
We defer the proof to \Cref{proof:unbiased}. REINFORCE with baseline is a special form of \Cref{eq:grad} when $n=m=1$.  
  
Beyond this basic formulation, several practical algorithms have been developed to improve training stability. PPO (Proximal Policy Optimization) \citep{schulman2017proximal} is widely used in RLHF pipelines for language models due to its clipping objective, which prevents overly large policy updates. It is defined as by the following updates, where we leave the advantage estimator  $A_{i,j,t}$ to be specified.
\begin{gather*}
\textstyle \mathcal{J}_{\mathrm{PPO}}(\theta) = \tfrac{1}{nm}\sum_{i,j}\tfrac{1}{|y_i^j|}\sum_{t=1}^{|y_i^j|} \min\bigl(\rho_t A_t,\,\mathrm{clip}(\rho_t,1\pm\epsilon)\,A_t\bigr),\\
\textstyle \rho_t = \tfrac{\pi_\theta(y_{i,t}^j\mid x_i,y_{i,<t}^j)}{\pi_{\theta_{\mathrm{old}}}(y_{i,t}^j\mid x_i,y_{i,<t}^j)}.
\end{gather*}
For LLM training, GRPO \citep{shao2024deepseekmath} proposes a Z-normalized advantage, where $\mu_i$ and $\sigma_i$ are the prompt-level mean and standard deviation of rewards, and $\delta > 0$ avoids division by zero:
\begin{gather*}
\textstyle A_{i,j}^{\mathrm{GRPO}} = (r_i^j - \mu_i)/(\sigma_i + \delta),\\
\textstyle \mu_i = \tfrac{1}{m}\sum_{j=1}^m r_i^j, \quad \sigma_i = \sqrt{\tfrac{1}{m}\sum_{j=1}^m (r_i^j - \mu_i)^2}.
\end{gather*}
More recent work has shown that dividing by $\sigma_i$ biases the objective \citep{liu2025understanding} without improving empirical performance, and can be omitted \citep{deepseekai2025deepseekv32pushingfrontieropen,khatri2025art}.

\vspace{-5pt}
\section{Derivation of the Method}
  
  \subsection{From Policy Gradient Variance to Value Function Estimators}
  \label{sec:pg_to_value}
  
  The purpose of a reinforcement learning baseline is to act as a control variate and thereby reduce the variance of the policy gradient estimator. 
  The baseline $b$ should be chosen so as to minimize the variance of the gradient estimator in \cref{eq:grad}. Since $g(\vec x, \vec Y; \theta)$ is vector-valued, its variance is naturally represented by the covariance matrix $\mathrm{Var}[g(\vec x, \vec Y;\theta)]$. A common scalar summary is the trace of this matrix, i.e., the sum of coordinate-wise variances, which is equivalent to the mean-squared error of the estimator:
\begin{align}
\label{eqn:varG}
\textstyle \mathrm{Var}[g] &\textstyle := \mathrm{Tr}(\mathrm{Var}[g(\vec x, \vec Y;\theta)]) \nonumber \\
&\textstyle = \E[\|g(\vec x, \vec Y;\theta) - \nabla_\theta J(\theta)\|_2^2].
\end{align}
In general, the variance-minimizing baseline for a given prompt $x$ depends on the squared norm of the score function $\nabla_\theta \log \pi_\theta(y \mid x)$, which is typically expensive to compute \citep{greensmith2004variance}. For this reason, a standard simplification in the literature---often made implicitly---is to ignore the dependence on the score function and directly find a baseline function $b(x)$ that minimizes the  Mean Square Error of the baseline estimator with respect to the observed rewards. In other words, theoretically one would choose a baseline $b$ to minimize the population-level mean square error for every state $x$:
\begin{align}
\label{eqn:popMSE}
\textstyle \mu(x)= \argmin_b \E[(r(x,y)-b(x))^2].
\end{align}
The optimal baseline is the value function, i.e., the \emph{Bayes-optimal predictor} of the reward under the current policy:
\[
\textstyle \mu(x) := \E_{y \sim \pi_\theta(\cdot \mid x)}[r(x,y)], \quad \forall x.
\]
The value-function baseline has become the standard foundation for variance reduction in reinforcement learning \citep{SuttonBarto2018-AC}, serving as the core theoretical basis for both classical actor--critic methods (e.g., A3C \citep{mnih2016asynchronous}, SAC \citep{haarnoja2018soft}, TRPO \citep{schulman2015trust}, PPO \citep{schulman2017proximal}) and more recent critic-free methods for reasoning models (e.g., ReMax \citep{li2023remax}, RLOO \citep{ahmadian2024back}, GRPO \citep{shao2024deepseekmath}, REINFORCE++ \citep{hu2025reinforce++}). 
  
  In practice, the value function is unknown. Simple algebra shows that minimizing \cref{eqn:popMSE}  leads to \textbf{minimizing the mean squared error of the baseline $b(x)$ with respect to the value function $\mu(x)$}, which is the starting point for our development to follow.
\begin{flalign}
&\textstyle \E[(r(x,y)-b(x))^2] = \E[(r(x,y)-\mu(x) + \mu(x) - b(x))^2] & \nonumber \\
&\textstyle = \E[(r(x,y)-\mu(x))^2] - 2\E[(r(x,y)-\mu(x))](b(x)-\mu(x)) & \nonumber \\
&\textstyle \quad + (b(x)-\mu(x))^2 & \nonumber \\
&\textstyle = \Var[r(x,y)] + (b(x)-\mu(x))^2. &
\label{eqn:startMSE}
\end{flalign}

\vspace{-3pt}
\subsection{The Bias-Variance Trade-off for Baselines in RLVR}
\label{sec:bias-variance-tradeoff}

We then analyze a single RL step with a batch of prompts. Corresponding to \cref{eq:grad}, policy-gradient updates also require value-function estimates for a \emph{batch} of prompts. This means that the empirical optimization problem associated with \cref{eqn:startMSE} is a batch-level mean square error:
\begin{equation}
\label{eq:mse_objective}
\textstyle \frac{1}{mn}\sum_{i=1}^n \sum_{j=1}^m \E_{\vec Y}\!\big[(b_i^j - \mu_i)^2\big].
\end{equation}
There is a standard bias–variance decomposition of the baseline mean-squared error. 
\begin{align}
& \textstyle \tfrac{1}{mn}\sum_{i=1}^n \sum_{j=1}^m \E[(b_i^j - \mu_i)^2] \label{eq:mse} \\
\textstyle = & \textstyle \tfrac{1}{mn}\sum_{i,j}\E[(b_i^j-\E[b_i^j]+\E[b_i^j]-\mu_i)^2] \notag \\
\textstyle = & \textstyle \tfrac{1}{mn}\sum_{i,j}(\E[(b_i^j-\E[b_i^j])^2] + 2\E[(b_i^j-\E[b_i^j])](\E[b_i^j]-\mu_i) \notag \\
& \textstyle \qquad\qquad + (\E[b_i^j]-\mu_i)^2) \notag \\
\textstyle = & \textstyle \underbrace{\tfrac{1}{mn}\sum_{i,j} \E[(b_i^j-\E[b_i^j])^2]}_{\text{Variance}} + \underbrace{\tfrac{1}{mn}\sum_{i,j}(\E[b_i^j]-\mu_i)^2}_{\text{Bias$^2$}}. \notag
\end{align}
Here expectations are taken over the responses $\vec Y$, and $\mu_i=\E_{y\sim\pi_\theta(\cdot|x_i)}[r(x_i,y)]$ denotes the true value function for prompt $x_i$. For exposition, we temporarily relax the unbiased-gradient requirement that $b_i^j$ be independent of $y_i^j$. 
  
\begin{figure*}[t]
    \centering
    \includegraphics[width=1.0\textwidth]{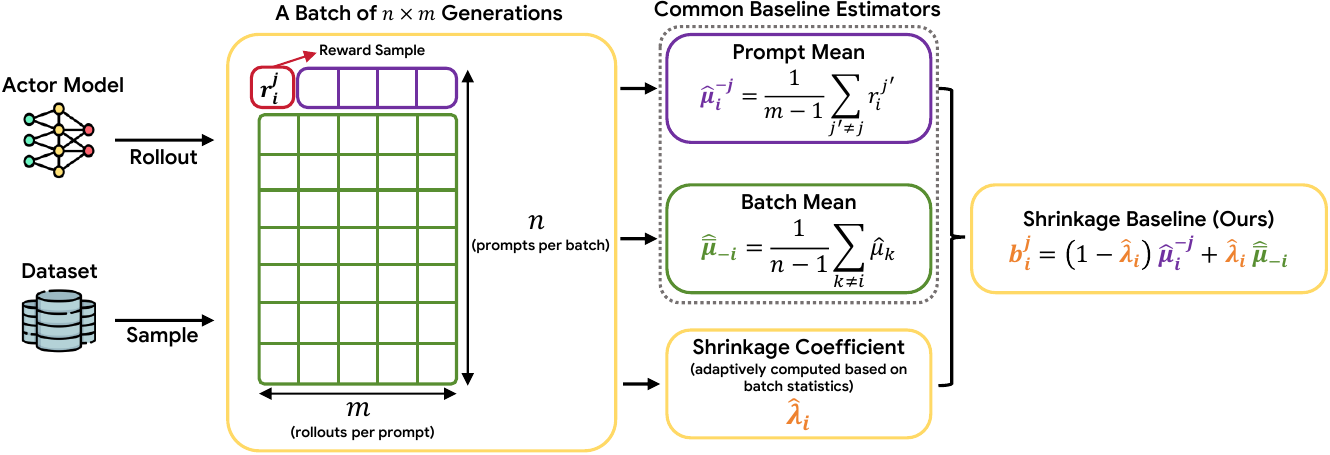}
    \captionsetup{skip=5pt}
    \caption{\textbf{Overview of using the Shrinkage Baseline in RLVR of large reasoning models.} 
    Consider a step in RLVR with $\textbf{n}$ question prompts, each generating $\textbf{m}$ responses. For every response, our method computes a leave-one-out prompt-level reward mean $\hat \mu_i^{-j}$ and a leave-one-out batch-level reward mean $\hat {\bar \mu}_{-i}$. It then estimates an optimal shrinkage coefficient $\hat \lambda_i$ from reward-sample statistics. These components are combined to produce a variance-reduced baseline $b_i^j$. By lowering the variance in policy-gradient estimation, the shrinkage baseline enables more stable and effective reinforcement learning for large reasoning models.}
    \label{fig:method}
\end{figure*}

The decomposition above highlights a bias–variance trade-off, with different baselines exhibiting distinct behaviors:
\begin{itemize}[leftmargin=*]
  \item \textbf{Prompt-level mean reward}, the most commonly used baseline in current RLVR, e.g., in RLOO \citep{ahmadian2024back} and GRPO \citep{guo2025deepseek}:
  \begin{equation}
    \label{eq:prompt_mean}
    \boxed{\hat \mu_i = \frac{1}{m}\sum_{j=1}^m r_i^j.}
  \end{equation}
  This choice minimizes the bias term, but does not efficiently reduce variance because it ignores informative cross-prompt structure.
  
  \item \textbf{Global batch mean reward}, as in REINFORCE++ \citep{hu2025reinforce++} and SPO \citep{xu2025single}:
  \begin{equation}
    \label{eq:global_mean}
    \boxed{\hat{\bar\mu} = \frac{1}{nm}\sum_{i=1}^n \sum_{j=1}^m r_i^j.}
  \end{equation}
  This baseline often achieves low variance but incurs substantial bias when prompt-specific means differ.
\end{itemize}
Given the limitations of both baselines, it is natural to consider combining them. While such a combination has not been studied in RL training, a similar mathematical question exists in statistics. In the classical multivariate normal setting, James \citep{james1961estimation} and Stein \citep{stein1956inadmissibility} showed that when estimating multiple means jointly, the empirical mean is provably suboptimal as long as there are more than three estimates, and a certain form of combination that pool information across tasks can strictly reduce total MSE. The reward distribution in RLVR is not Gaussian, so the classical James--Stein guarantees do not directly apply. Nevertheless, the underlying principle of borrowing strength across related estimation problems remains appealing in our setting. This motivates us to consider shrinking each prompt-level mean toward the global batch mean, yielding a \emph{shrinkage baseline} in RLVR:
\begin{align}
\label{eqn:JSnaive}
  \boxed{b_i^{j,\mathrm{JS1}}
  = (1-\lambda)\,\hat\mu_i + \lambda\,\hat{\bar\mu}.}
\end{align}
  The shrinkage coefficient $\lambda\in[0,1]$ balances variance reduction and bias.
  The optimal $\lambda$ admits a closed-form estimate from data (proof in Appendix~\ref{proof:mse_motiv}):
  \begin{propositionbox}
  \begin{proposition}
  \label{lem:mse_motiv}
  Let $v=\tfrac{1}{nm}\sum_{i=1}^n\sigma_i^2$, where $\sigma_i^2=\mathrm{Var}[r(x_i,y)]$. Let $\bar \mu=\tfrac{1}{n}\sum_{i=1}^n \mu_i$ and $s=\tfrac{1}{n-1}\sum_{i=1}^n(\mu_i-\bar \mu)^2$. Then the minimizer of the relaxed MSE is
\begin{align*}
\lambda^\star \;=\; \tfrac{v}{s+v}.
\end{align*}
\end{proposition}
\end{propositionbox}
Here, $v$ captures the average per-prompt variance, while $s$ measures the dispersion of true value functions across prompts. When rewards of prompts are similar (small $s$), stronger shrinkage is preferred; when rewards of prompts are heterogeneous (large $s$), the estimator leans more on local means. \textbf{Since $0<\lambda^\star<1$ whenever $v>0$ and $s>0$, both the prompt-level mean reward in \Cref{eq:prompt_mean} and the global batch mean reward in \Cref{eq:global_mean} are actually strictly suboptimal for variance reduction}. Instead, the baseline that interpolates them together as \Cref{eqn:JSnaive} is actually the optimal balance between \emph{batch-level} and \emph{prompt-level} information, which \textbf{dominates both empirical means} under the objective formulated by \Cref{eq:mse_objective}. An overview of our method is shown in \Cref{fig:method}.

\subsection{RLVR with Shrinkage Baseline}
  
  The analysis in the prior section suggests that a shrinkage baseline with lower MSE can substantially reduce policy gradient variance. However, a critical additional requirement in policy gradient is \emph{unbiasedness}. The naive shrinkage baseline in \Cref{eqn:JSnaive} cannot be used directly, since it is correlated with the rewards $r_i^j$ that appear in the gradient estimator, and thus would introduce bias. 
  
  In this section, we consider the general case that prompts are sampled from $\cD$ instead of being fixed, and derive a more general shrinkage baseline with unbiasedness guarantee. To guarantee independence between the baseline and each individual reward, we adopt a two-level leave-one-out construction in the spirit of RLOO. For each prompt $x_i$ and response $y_i^j$, define
\begin{gather}
    \textstyle \hat \mu_i^{-j} := \frac{1}{m-1}\sum_{j'\neq j} r_i^{j'}, 
    \textstyle \hat{\bar \mu}_{-i} := \frac{1}{n-1}\sum_{k\neq i}\hat \mu_k, \label{eq:local_loo} \\
    \textstyle b_i^{j,\mathrm{JS2}} := (1-\lambda_i^{j})\,\hat \mu_i^{-j} + \lambda_i^j \,\hat{\bar \mu}_{-i}.
    \label{eq:baseline_unbias}
\end{gather}
  Compared to \Cref{eqn:JSnaive}, \Cref{eq:baseline_unbias} replaces both the prompt-level and batch-level means with leave-one-out counterparts, ensuring that $b_i^{j,\mathrm{JS2}}$ is independent of the held-out reward $r_i^j$. Furthermore, we allow the shrinkage coefficient to vary by sample, i.e., each $b_i^j$ uses its own $\lambda_i^j$, which can also be chosen independently of $r_i^j$. With these modifications, the resulting estimator yields an unbiased policy gradient. 
  
  The optimal shrinkage coefficient has essentially the same structure as in the naive shrinkage baseline estimator. Define $\mu(x):=\E_{y\sim\pi(\cdot\mid x)}[r(x,y)]$ and
$\sigma^2(x):=\Var_{y\sim\pi(\cdot\mid x)}[r(x,y)]$.
  The following theorem provides its precise expression.
  
  \begin{theorembox}
  \begin{theorem}
  \label{lem:mse_unbias}
  Let $v_2=\textstyle \tfrac{1}{m-1}\E_{x\sim\mathcal D}[\sigma^2(x)]$ and
  $s_2=\Var_{x\sim\mathcal D}[\mu(x)]$.
   Then the optimal shrinkage coefficient for \Cref{eq:mse} is the same across all prompts $i$ and samples $j$, and is given by
  \begin{align}
  (\lambda_i^j)^* \;=\; \tfrac{n-1}{n}\cdot \tfrac{v_2}{s_2 + v_2}.
  \label{eq:js-opt-coeff}
  \end{align}
  \end{theorem}
  \end{theorembox}
  Here $v_2 = \textstyle \tfrac{1}{m-1} \E_{x \sim \mathcal{D}}[\sigma^2(x)]$ measures the \emph{expected variance} of the estimator $\hat{\mu}_i^{-j}$, while $s_2 = \textstyle \mathrm{Var}_{x \sim \mathcal{D}}[\mu(x)]$ quantifies the variability of the true value functions across prompts. In Appendix~\ref{proof:mse_unbias}, we provide a proof of \Cref{lem:mse_unbias}. Notably, \Cref{lem:mse_unbias} does not rely on any parametric assumptions on the joint distribution of prompts and rewards.
  
  \begin{itemize}[leftmargin=*, topsep=2pt, itemsep=2pt]
      \item When $v_2 \gg s_2$, the \textbf{per-prompt estimates are highly noisy}. This typically occurs when only few rollouts per prompt are available, and in such cases, \textbf{our shrinkage baseline naturally degrades to the global batch mean baseline}.
      \item When $s_2 \gg v_2$, the \textbf{value functions vary substantially across prompts}. In such cases, the optimal $\lambda$ is close to zero, and \textbf{our shrinkage baseline naturally degrades to the prompt-level mean baseline}.
  \end{itemize}
Thus, $(\lambda_i^j)^*$ achieves the optimal trade-off between variance reduction and bias control, while preserving the independence condition required for unbiased policy gradients. 
  
\paragraph{Implementation details}
  In practice, we first estimate $\widehat{v}_{-i}$ and $\widehat{s}_{-i}$ from statistics among batch leave-one-out reward samples:
\begin{align}
\widehat{v}_{-i} &= \textstyle \tfrac{1}{n-1} \sum_{k\ne i}
        \left(
            \tfrac{1}{m(m-1)} \sum_{j=1}^{m} (r_k^j - \hat\mu_k)^2
        \right), \label{eq:v-est} \\
\widehat{s}_{-i} &= \textstyle \tfrac{1}{n-1} \sum_{k\ne i}
        (\hat\mu_k - \hat{\bar\mu}_{-i})^2. \label{eq:s-est}
\end{align}
  Given these plug-in estimates, the per-prompt shrinkage coefficient becomes
\begin{align}
\hat \lambda_i = \textstyle \tfrac{n-1}{n} \cdot
    \tfrac{\widehat{v}_{-i}}{\widehat{v}_{-i} + \widehat{s}_{-i}}.
\label{eq:lambda-est}
\end{align}
  Finally, combining the local and global averages (\Cref{eq:local_loo}) with the estimated $\hat \lambda_i$ in \Cref{eq:lambda-est} yields the \emph{shrinkage baseline}:
\begin{align}
b_i^j = \textstyle (1 - \hat \lambda_i)\, \hat \mu_i^{-j} + \hat \lambda_i\, \hat{\bar \mu}_{-i}.
\label{eq:js-final-baseline}
\end{align}
  It is worth noting that $b_i^j$ here does not depend on the response $y_i^j$, so it satisfies the condition of \Cref{prop:unbiased}, and thus the resulting gradient is \textbf{unbiased}.

\section{Experimental Analysis}
  \label{sec:exp}
  
  In this section, we empirically evaluate the effectiveness of the proposed shrinkage baseline (also named \textbf{J}ames-\textbf{S}tein baseline) in reinforcement finetuning of large reasoning models. Through experimental analysis, we show that shrinkage baseline benefits reinforcement learning by reducing the variance of the policy gradient. For the main experimental setup, we adopt the GRPO algorithm \citep{shao2024deepseekmath} without advantage normalization as recommended by a recent large-scale empirical investigation \citep{khatri2025art}, and with leave-one-out reward centering, also known as RLOO \citep{ahmadian2024back}. 

  \vspace{-8pt}

  \subsection{Mathematical Reasoning}
  \label{sec:exp-math}

  \begin{figure}[htbp]
    \centering
    \includegraphics[width=0.9\linewidth]{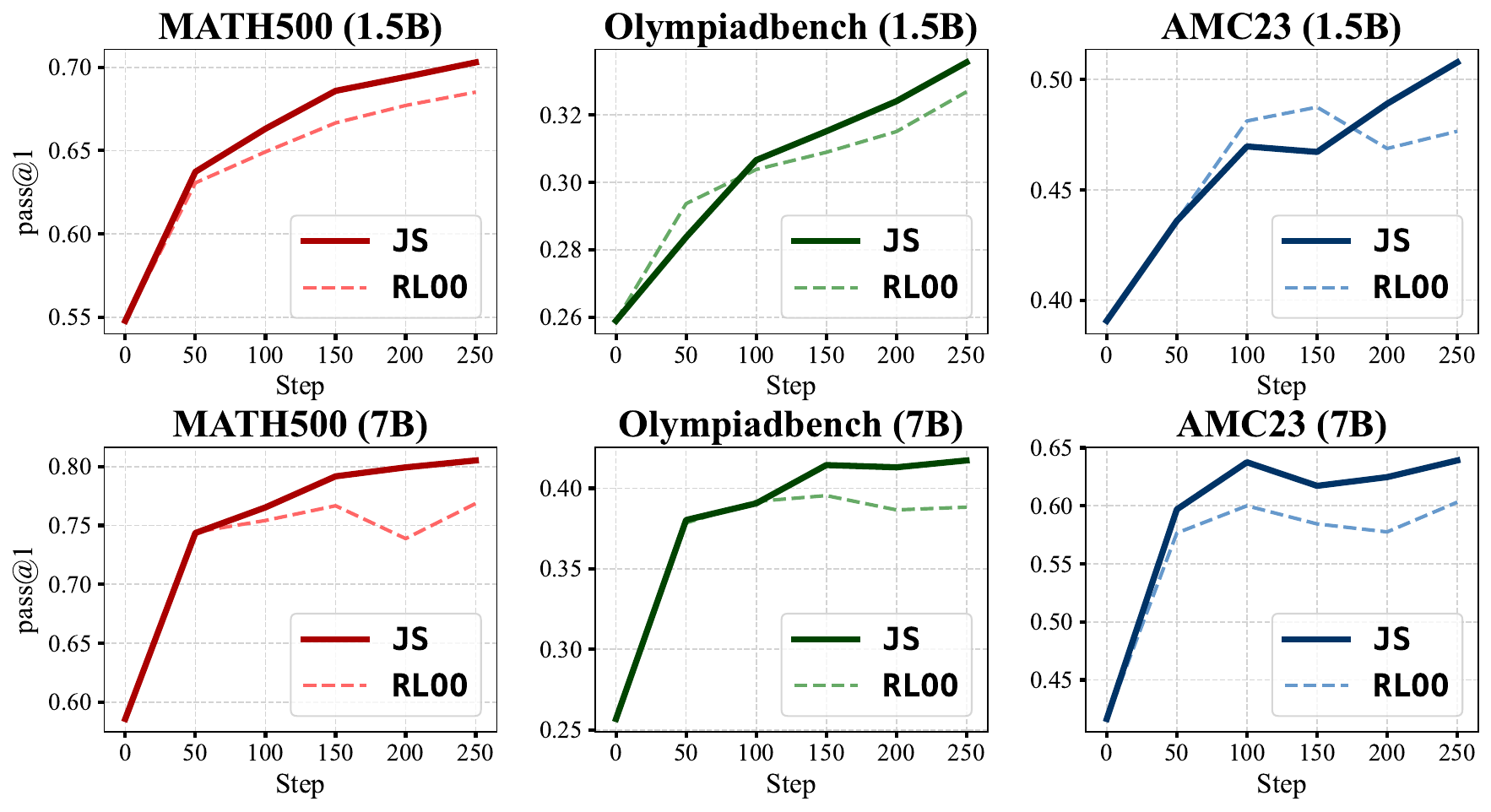}
    \captionsetup{skip=15pt}
    \caption{Comparison of shrinkage baseline with RLOO \citep{ahmadian2024back} baseline on Qwen2.5 math models trained on DAPO17k dataset. shrinkage baseline significantly outperforms RLOO across different models and benchmarks.}
    \vspace{-15pt}
    \label{fig:math}
  \end{figure}
  
  We first evaluate the shrinkage baseline on mathematical reasoning tasks. In this section, we use Qwen2.5-Math-1.5B~\citep{qwen2.5}, Qwen2.5-Math-7B~\citep{qwen2.5}, and Qwen3-4B-Base~\citep{yang2025qwen3} as base models and train on the DAPO17k~\citep{yu2025dapo} dataset. During training, we use 64 questions per batch with 4 rollouts per question. For the Qwen2.5 math models, we set the maximum token length to 2048 and evaluate on three commonly used benchmarks: MATH500~\citep{hendrycks2021measuring}, OlympiadBench~\citep{he2024olympiadbench}, and AMC23~\citep{amc23_2025_hf}. For the Qwen3-4B-Base model, we extend the maximum token length to 3072, increase the clip ratio to 0.25, and adopt the length-dependent loss aggregation technique as suggested by DAPO~\citep{yu2025dapo}.
  
\begin{figure}[htbp]
  \centering
  \includegraphics[width=0.8\linewidth]{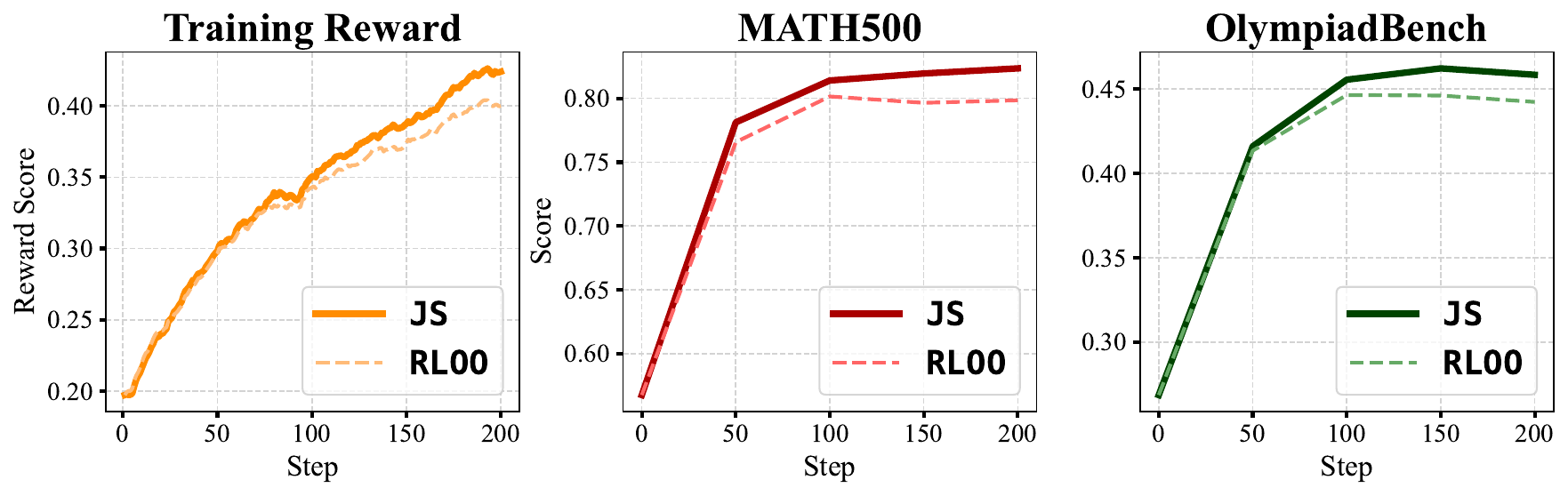}
  \captionsetup{skip=15pt}
  \caption{Comparison of training reward and test accuracy between shrinkage baseline and RLOO on Qwen3-4B-Base model trained on DAPO17k dataset.}
  \label{fig:math2}
\end{figure}

  \Cref{fig:math} shows the accuracy of math reasoning on Qwen2.5 math models, measured by the average Pass@1 over 16 samples, and \Cref{fig:math2} presents the training reward curve and test accuracy of the Qwen3-4B-Base model. Compared to the RLOO baseline~\citep{ahmadian2024back}, which uses only the leave-one-out mean reward within each question without shrinkage toward the batch mean, our shrinkage baseline achieves substantial improvements across different models and benchmarks, yielding \textbf{2.3\%--5.5\% relative gains in Pass@1} along with significantly faster reward growth during training. Additional experimental details and training results are provided in Appendix~\ref{app:detail-math}.
  
\vspace{-5pt}
  \subsection{Logic Puzzle Reasoning}
  \label{sec:exp-logic}

  We then apply the shrinkage baseline into various logic puzzle reasoning tasks. Following prior work~\citep{tinyzero,chen2025pass,stojanovski2025reasoning}, we adopt three settings that are suitable for reinforcement finetuning in terms of model capability and task difficulty: Qwen2.5-7B-Instruct~\citep{qwen2.5} on Knights-and-Knaves (KnK)~\citep{stojanovski2025reasoning}, Qwen2.5-3B~\citep{qwen2.5} on Countdown~\citep{gandhi2024stream, tinyzero}, and Ministral-8B-Instruct~\citep{mistral7b} on Maze~\citep{chen2025pass}. For each model, we train for at least 200 steps and evaluate on a held-out test set of at least 200 problems drawn from the same distribution as the training data. Additionally, we train Qwen2.5-1.5B-Instruct~\citep{qwen2.5} on the KnK dataset for up to 1000 steps with a smaller learning rate to further validate the effectiveness of our method. Detailed descriptions of each puzzle, along with additional experimental results, are provided in Appendix~\ref{app:detail-logic}.

  \begin{figure}[htbp]
    \centering
    \includegraphics[width=0.7\linewidth]{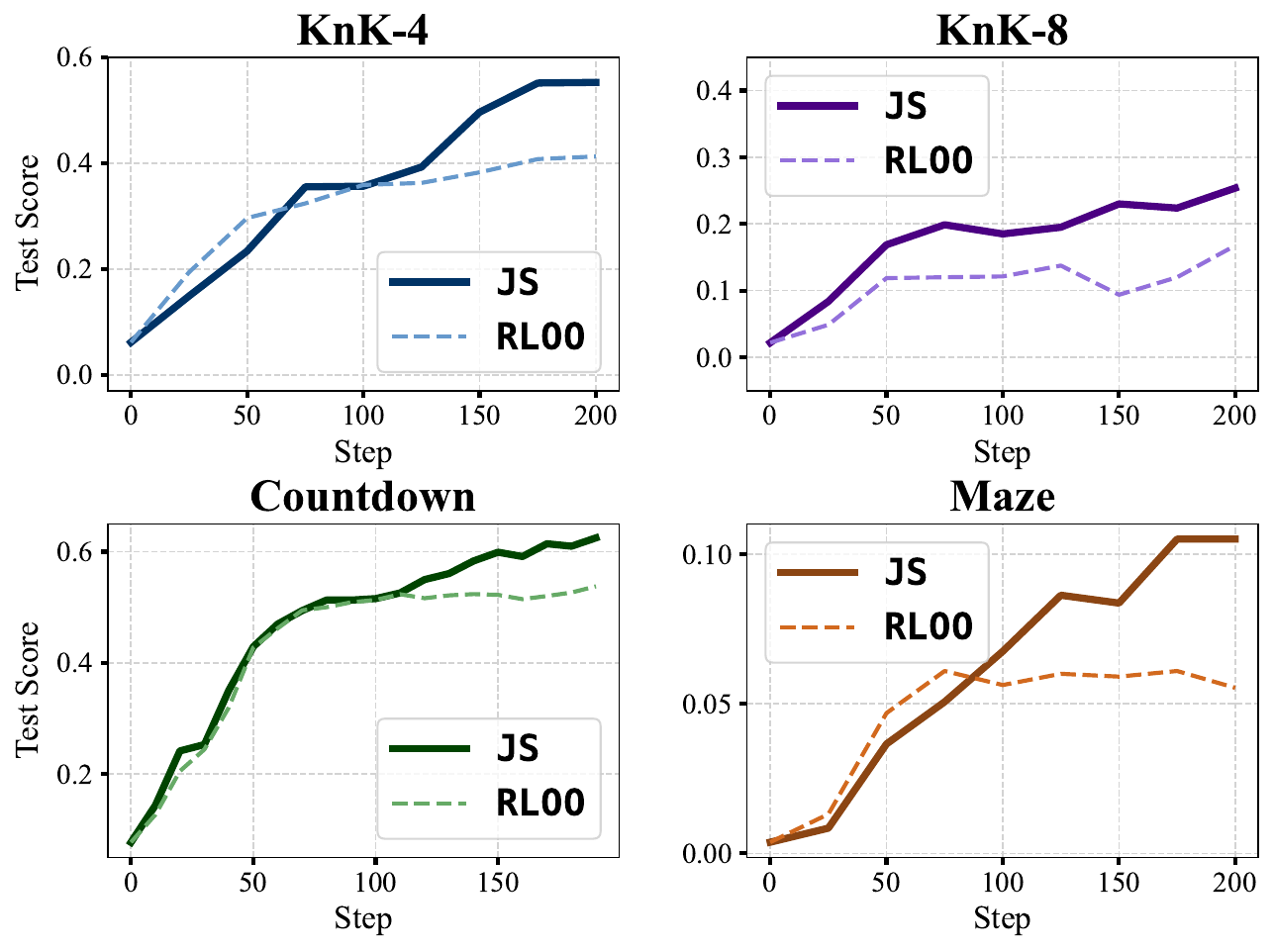}
    \caption{Comparison of average test scores between the shrinkage baseline and RLOO on logic puzzle reasoning tasks. The shrinkage baseline outperforms RLOO across all tasks and models. The number following Knights-and-Knaves (KnK) indicates the number of people in the puzzle; larger numbers correspond to higher difficulty.}
    \label{fig:logic}
  \end{figure}

\begin{figure}[htbp]
  \centering
  \includegraphics[width=0.9\linewidth]{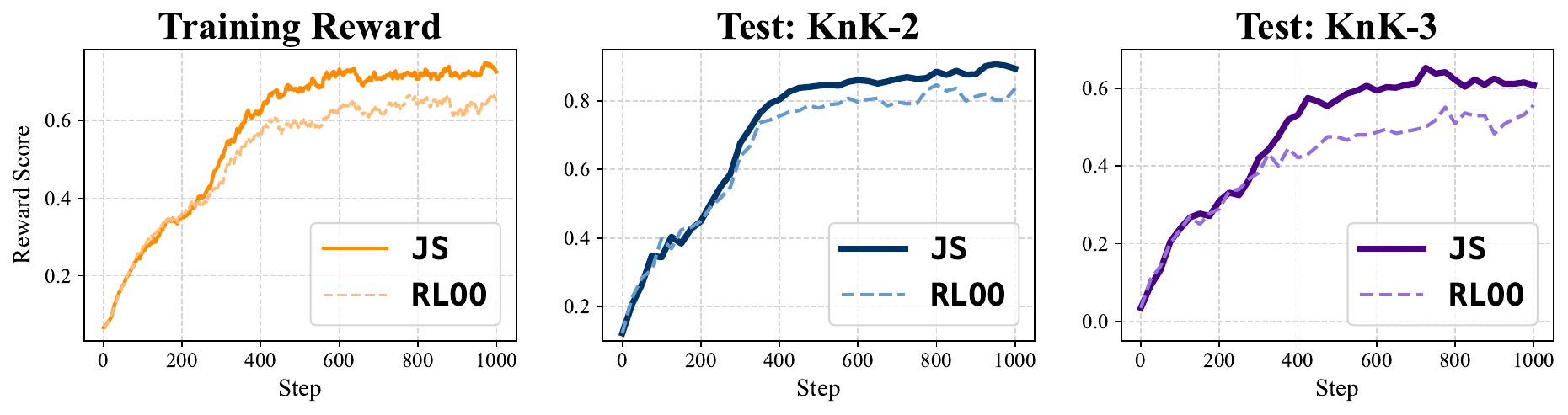}
  \caption{Comparison of training reward (running average) and test accuracy between the shrinkage baseline and RLOO on the Qwen2.5-1.5B-Instruct model trained on the KnK dataset.}
  \label{fig:logic2}
\end{figure}
  
  \Cref{fig:logic} plots the average test scores over training steps across different logic puzzles, and \Cref{fig:logic2} shows the training reward and test accuracy of the Qwen2.5-1.5B-Instruct model over extended training. Compared to the RLOO baseline, the shrinkage baseline yields substantial improvements, achieving \textbf{4.8\%--15.2\% gains in Pass@1} on the test sets. These results demonstrate that the shrinkage baseline enables more stable and effective parameter updates during reinforcement learning of reasoning LLMs. 

  \subsection{Multimodal Reasoning}
  \label{sec:exp-multimodal}
  
\begin{figure}[htbp]
  \centering
  \includegraphics[width=0.8\linewidth]{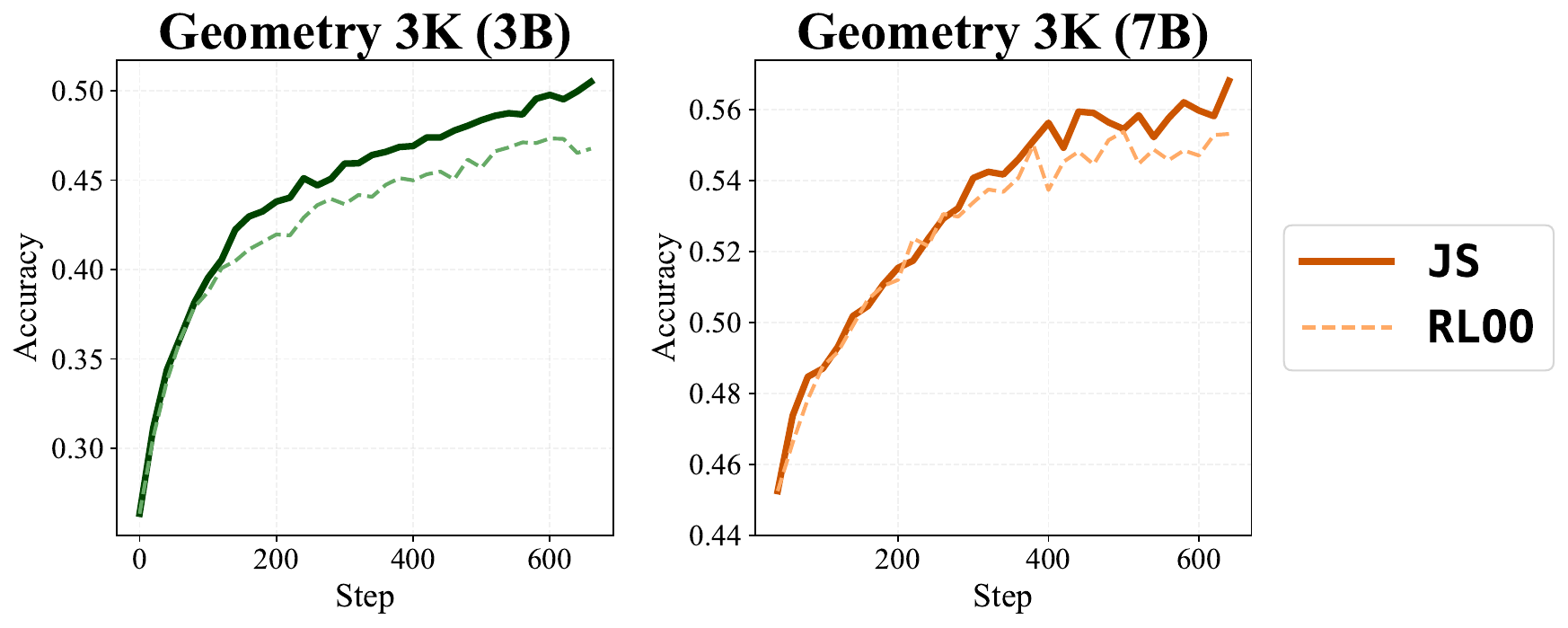}
  \caption{Comparison of test accuracy between the shrinkage baseline and RLOO on Qwen2.5-VL-3B-Instruct and Qwen2.5-VL-7B-Instruct trained on Geometry3k dataset.}
  \vspace{-8pt}
  \label{fig:geo3k}
\end{figure}

  We also leverage the shrinkage baseline for visual reasoning tasks. Following the training recipe in verl~\citep{verl2024github}, we train Qwen2.5-VL-3B-Instruct~\citep{bai2025qwen25vltechnicalreport} and Qwen2.5-VL-7B-Instruct~\citep{bai2025qwen25vltechnicalreport} on the Geometry3k~\citep{lu2021intergps} dataset for 600 gradient steps. This dataset contains 2,101 training examples of challenging geometry questions which requires a mixed understanding and reasoning on the basis of both text and image information. During training, we evaluate on the test split of Geometry3k which consists 601 questions for every 20 gradient steps. As shown in \Cref{fig:geo3k}, the shrinkage baseline consistently outperforms RLOO baseline throughout training, yielding a \textbf{4.4\% and 3.8\% relative gain in final test accuracy}. Additional details are provided in Appendix~\ref{app:multimodal}.

  \subsection{Reinforcement Learning with Human Feedback}

  Beyond reasoning tasks, we further extend our shrinkage baseline to another important LLM post-training setting: reinforcement learning from human feedback (RLHF). The goal of RLHF is to align an LLM's outputs with human preferences, which are usually modeled by a Bradley–Terry reward model, instead of binary rewards in previous three experiments. Following previous work \citep{hu2025reinforce++,tiapkin2025accelerating,zhang2025onpo,leng2024taming}, we use a comprehensive training dataset \citep{dong2024rlhfworkflow} that contains 179k prompts from six widely used RLHF datasets: UltraFeedback \citep{cui2024ultrafeedback}, HelpSteer \citep{wang2024helpsteer}, OpenOrca \citep{openorca2023}, UltraInteract \citep{yuan2025ultrainteract}, DIBT-10K \citep{dibt2024_10k_prompts_ranked}, and Capybara Preferences \citep{capybara_preferences_2024}. We train a Llama-3.2-3B-Instruct model \citep{grattafiori2024llama} for one epoch using an 8B reward model \citep{openrlhf_llama3_8b_rm_700k} trained on 700k preference pairs, and evaluate on three widely used benchmarks: Arena-Hard-v0.1 \citep{li2024crowdsourced_v2}, Arena-Hard-v2.0 \citep{li2024crowdsourced_v2}, and Arena-Creative-Writing \citep{li2024crowdsourced_v2}. As shown in \Cref{tab:arena-results}, the shrinkage baseline consistently outperforms the RLOO baseline on all three benchmarks. These results show that the effectiveness of shrinkage baseline is not confined to rule-based, binary reward tasks.

\begin{table}[htbp]
  \caption{RLHF benchmark results (win rates). AH = Arena-Hard, CW = Arena-Creative-Writing. Green values indicate relative improvement over RLOO.}
  \small
  \label{tab:arena-results}
  \centering
  \begin{tabular}{lccc}
    \toprule
    \textbf{Method} & \textbf{AH-v0.1} & \textbf{AH-v2.0} & \textbf{CW} \\
    \midrule
    Llama-3.2-3B-It & 26.2\% &  2.4\% &  5.2\% \\
    RLOO            & 55.3\% &  5.6\% & 20.2\% \\
    JS              & \textbf{56.7\%} {\scriptsize\textcolor[rgb]{0,0.5,0}{+2.5\%}} &
                      \textbf{7.4\%} {\scriptsize\textcolor[rgb]{0,0.5,0}{+32.1\%}} &
                      \textbf{23.4\%} {\scriptsize\textcolor[rgb]{0,0.5,0}{+15.8\%}} \\
    \bottomrule
  \end{tabular}
\end{table}

  \subsection{Comparison with Other Baselines}
  \label{sec:exp-comparison}
  
  We explore the performance of shrinkage baseline under different number of rollouts and systematically compare with other variance reduction baselines. For computation efficiency, we train Qwen2.5-0.5B-Instruct \citep{qwen2.5} model on GSM8k \citep{cobbe2021training} dataset for 500 steps. For each RL step, we sample 64 questions, and vary the number of rollouts (i.e. number of generations per question) among 2,4,8. We experiment on different critic-free RLVR baselines, including GRPO baseline \citep{shao2024deepseekmath}, RLOO baseline \citep{ahmadian2024back}, ReMax baseline \citep{li2023remax}, REINFORCE++ baseline \citep{hu2025reinforce++}, batch-level leave one out (BLOO) and JS baseline. BLOO means computing the baseline by the average of rewards within the batch with the current prompt left out, i.e., it uses $\hat{\bar \mu}_{-i}$ in \Cref{eq:js-final-baseline}. For each setting, we iterate over 5 random seeds and report the average final accuracy on the GSM8k test split.

  \begin{table}[htbp]
  \centering
  \footnotesize
  \caption{\textbf{Final test accuracy (\%) across various number of rollouts.} 
  Best are in bold, second-best with *, third-best with \textsuperscript{\dag}.}
  \label{tab:final-accuracy-2}
  \begin{tabular}{@{}lccc@{}}
    \toprule
    Baseline & 2 Gen & 4 Gen & 8 Gen \\
    \midrule
    ReMax        & 54.70\textsuperscript{\dag} & 56.47*  & 57.76 \\
    REINFORCE++  & 54.82*  & 55.27  & 57.30 \\
    GRPO         & 53.68  & 56.24  & 58.28\textsuperscript{\dag} \\
    BLOO         & 54.19  & 56.06  & 57.22 \\    
    RLOO         & 54.49  & 56.34\textsuperscript{\dag}  & 58.31* \\
    JS           & \textbf{55.22}  & \textbf{57.33}  & \textbf{58.93} \\
    \bottomrule
  \end{tabular}
\end{table}
    
  As shown in \Cref{tab:final-accuracy-2}, using the shrinkage baseline consistently achieves the highest evaluation scores across all rollout settings, whereas competing baselines only excel under specific conditions. RLOO and GRPO perform best with 8 rollouts, where prompt-level reward averaging becomes accurate, while REINFORCE++ and BLOO fare relatively better with only 2 rollouts, since GRPO and RLOO suffer from higher variance and batch-level averaging provides more stability. ReMax++ delivers modest performance across all settings, consistent with the limitations of its greedy decoding design. 

  \subsection{Analysis of Training Dynamics}
  \label{sec:exp-dyn}
  
  We focus on providing insights into two key training dynamics that reveal the advantage of using the JS baseline: namely, the \textit{Adaptive Shrinkage} of the James–Stein coefficient and the \textit{Reduced Variance} of the policy gradient. We provide more results illustrating these two effects in Appendix~\ref{app:supplementary-results}.
  
\begin{figure}[htbp]
    \centering
    \includegraphics[width=0.8\linewidth]{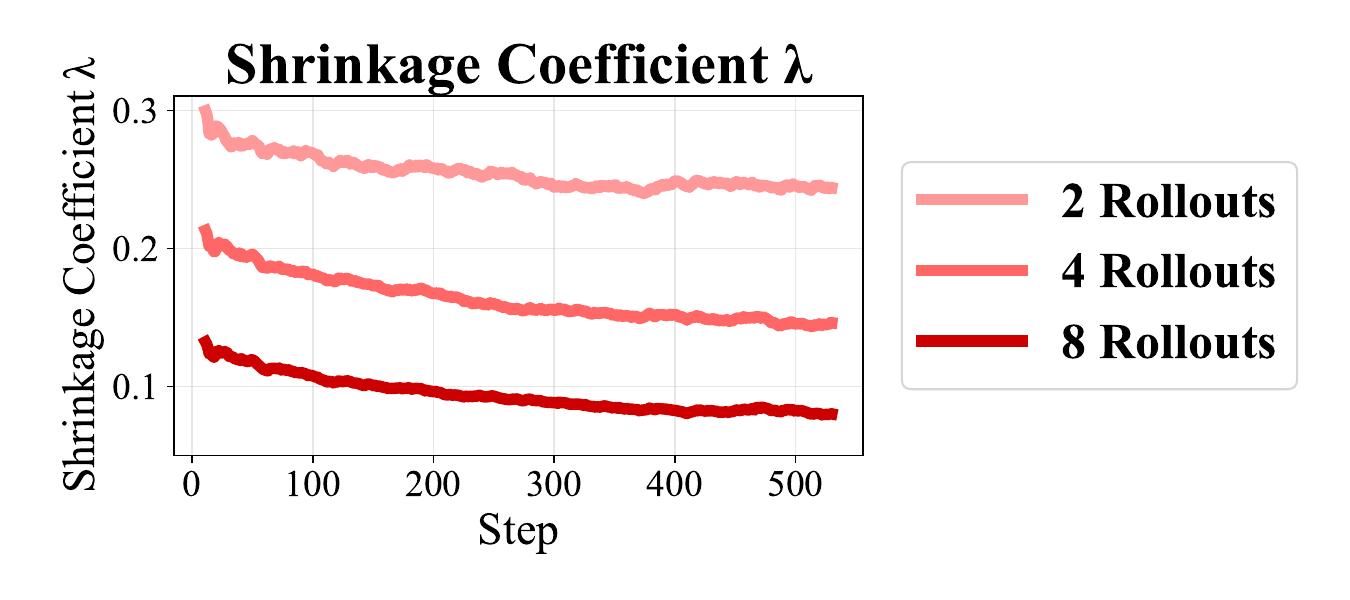}
    \caption{\textbf{Moving average shrinkage coefficient.} \(\hat{\lambda}_i\) during training at different numbers of rollouts.}
    \label{fig:js_lambda}
  \end{figure}

  \noindent \textbf{Adaptive Shrinkage.} The shrinkage coefficient $\hat \lambda_i$ in \Cref{eq:js-final-baseline} is central to JS baseline. \Cref{fig:js_lambda} reports its average value during training across different rollout counts, revealing two key trends: (i) with more rollouts, $\hat \lambda_i$ decreases, since intra-prompt estimates become more reliable and shrinkage baseline naturally approaches the RLOO baseline; and (ii) $\hat \lambda_i$ decays over training, as RLVR drives the policy toward greater determinism, reducing the usefulness of cross-prompt references. Together, these behaviors constitute an adaptive shrinkage mechanism that adjusts with both rollout number and training progress, explaining why shrinkage baseline consistently outperforms RLOO and BLOO across all rollout settings in \Cref{sec:exp-comparison}.

  \noindent \textbf{Reduced Variance.} The variance of the policy gradient in \Cref{eqn:varG} is the key metric that reflects the stability of the reinforcement finetuning. To track the gradient variance, we need to build an unbiased estimator of $\Var(g)$ using the observable gradients during training. Following \citet{mccandlish2018empirical}, we collect $m$ micro-batches of gradients $g_i$ ($i=1,\dots,m$) in one training step, then an unbiased estimator for $\Var(g)$ becomes
\begin{align}
\label{eq:unbiased_gradient_trace_cov}
\widehat{\Var(g)}
&= \textstyle \tfrac{1}{m}\cdot\tfrac{1}{m-1}\sum_{i=1}^m \|g_i-\bar g\|^2 \\
&= \textstyle \tfrac{1}{m}\cdot\tfrac{1}{m-1}\bigl(\sum_{i=1}^m \|g_i\|^2
- \tfrac{1}{m}\bigl\|\sum_{i=1}^m g_i\bigr\|^2\bigr)
\end{align}
  where $\bar g = \textstyle \tfrac{1}{m}\sum_{i=1}^m g_i$.
  We incorporate this estimator during training. Further implementation details are provided in \Cref{app:variance-recording}. 
  
\begin{figure}[htbp]
  \centering
  \includegraphics[width=0.85\linewidth]{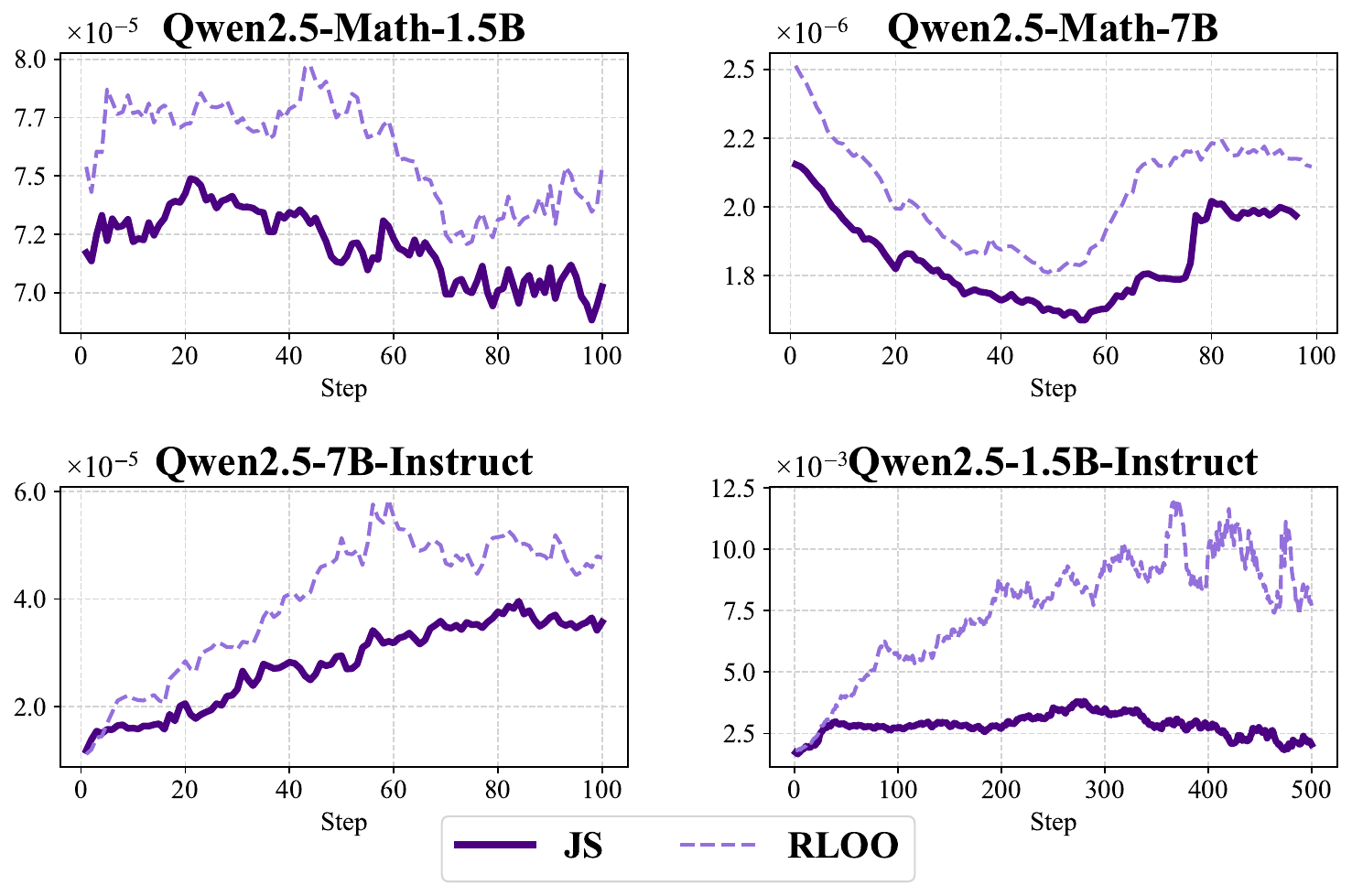}
  \caption{\textbf{Estimated variance of policy gradient during training in different models.} With baseline shrinkage, the gradient variance is significantly reduced.}
  \label{fig:var}
\end{figure}

  \Cref{fig:var} reports the running average of gradient variance across different experiments and models. Compared to RLOO, training with shrinkage baseline reduces gradient variance by 11.2\%, 17.4\%, 31.6\% and 67.1\%, respectively, suggesting a more stable training process.

To further measure the variance reduction mechanism, we estimate the ground-truth value function and policy gradient via extensive Monte Carlo sampling on saved checkpoints. As detailed in Appendix \ref{app:direct-verify}, the shrinkage baseline yields lower mean squared error with respect to the ground-truth value function (Appendix \ref{app:mse}) and produces gradient updates closer to the ground-truth policy gradient in terms of L2 norm (Appendix \ref{app:gt-pg}). In addition, we examine the robustness of the shrinkage baseline under varying training data distributions in Appendix \ref{app:ablation}. Results show that the shrinkage coefficient adapts to dataset difficulty heterogeneity while consistently outperforming the non-shrinkage baseline (Appendix \ref{app:data-heter}). Moreover, increased semantic heterogeneity primarily affects the magnitude of variance reduction, yet the shrinkage baseline remains optimal (Appendix \ref{app:semantic-heter}). Together, these results suggest that the shrinkage baseline provides robust variance reduction across diverse situations in LLM reinforcement learning.

\section{Related Work}
\label{sec:extended_related_work}

\subsection{Reinforcement Learning with Verifiable Rewards (RLVR)}

RLVR refers to a training paradigm where the reward is computed by a rule-based verification function—typically indicating whether the model's final answer is correct. This approach has proven effective in enhancing the reasoning capabilities of LLMs. The recent success of RLVR is closely tied to advances in reinforcement learning algorithms, which can be broadly categorized into two groups. \textbf{Actor-critic methods}, such as PPO \citep{schulman2017proximal} and its variants (e.g., VC-PPO \citep{yuan2025s}, VinePPO \citep{kazemnejad2024vineppo}, VAPO \citep{yuan2025vapo}), rely on training an additional value model (critic) to estimate baselines. While theoretically grounded, these methods incur high computational overhead. \textbf{Critic-free methods}, including RLOO \citep{ahmadian2024back}, ReMax \citep{li2023remax}, GRPO \citep{shao2024deepseekmath}, DAPO \citep{yu2025dapo}, Dr.GRPO \citep{liu2025understanding}, GSPO \citep{zheng2025groupsequencepolicyoptimization} and CISPO \citep{minimax2025minimaxm1scalingtesttimecompute} eliminate the need for a learned value function by directly estimating baselines or advantages from multiple responses to the same prompt. These methods significantly reduce training cost and have become the dominant approach in practical RLVR pipelines. Their effectiveness largely hinges on the quality of the estimated baseline, which serves as a variance reduction tool in policy gradient updates.

While RLVR inherits techniques from classic RL, a distinctive characteristic when applied to LLMs is the use of large batch sizes but small rollout numbers per prompt. For example, rollout numbers of 2, 4, or 8 are typically adopted~\citep{shao2024deepseekmath,verl_deepseek671b_grpo_recipe,openrlhf_llama31_8b_grpo_recipe,zhang2025r1vl,chen2025mintcot,kang-etal-2025-grpo,phan2025beyond_reasoning_gains}, and even large-scale industrial post-training practices~\citep{yang2025qwen3,guo2025deepseek,deepseekai2025deepseekv32pushingfrontieropen} rarely sample more than 32 responses per prompt due to the high cost of LLM inference. 

\subsection{Baselines in Policy Gradient Methods}

The use of baselines in policy gradient methods was originally introduced \citep{williams1992simple} as a variance reduction technique without introducing bias. Early work formalized this as a control variate problem, showing that the optimal constant baseline is the average return \citep{weaver2013optimal}, while state-dependent baselines can further reduce variance \citep{greensmith2004variance}. Actor-critic methods \citep{barto1989learning} extend this idea by learning value function approximations, and techniques such as generalized advantage estimation (GAE) \citep{schulman2015high} trade off bias and variance to improve stability and sample efficiency. More recent work explores state-action-dependent baselines: Q-Prop \citep{gu2016q} leverages off-policy critics as control variates, and action-dependent factorized baselines \citep{wu2018variance} exploit policy structure to achieve lower variance in high-dimensional settings. Other methods use Stein’s identity to learn expressive baselines with action dependency \citep{liu2017action}. However, later empirical study \citep{tucker2018mirage} suggest that when value functions are well-approximated, simple state-dependent baselines can match or outperform more complex alternatives. In summary, an effective baseline should minimize variance, maintain zero or low bias, and be robust to implementation, with growing consensus emphasizing careful design over complexity.

\subsection{Stein's Paradox}

Stein's paradox~\citep{stein1956inadmissibility} highlights a striking fact in multivariate decision theory: when estimating the mean vector $\theta \in \mathbb{R}^p$ from a single observation $X \sim \mathcal{N}_p(\theta, \sigma^2 I)$ under squared-error loss, the usual estimator $\hat{\theta}=X$ (equivalently, the coordinatewise MLE) is inadmissible as soon as $p\ge 3$. In contrast, James--Stein (JS) estimator~\citep{stein1956inadmissibility,james1961estimation} achieves uniformly smaller risk by shrinking the observation toward a common target, thereby ``borrowing strength'' across coordinates and reducing overall mean squared error. 
\begin{equation}
\label{eq:classical-js}
\hat{\theta}_{\mathrm{JS}}(X)
= \mu + \left(1 - \frac{(p-2)\sigma^2}{\|X-\mu\|_2^2}\right)(X-\mu)
\end{equation}
This phenomenon sparked applications of shrinkage rules and their decision-theoretic properties, including empirical Bayes interpretations and data-driven shrinkage \citep{efron1973stein}, minimax shrinkage families \citep{baranchik1970family}, and admissibility characterizations via connections to potential theory and diffusions \citep{brown1971admissible,berger1976admissible}. Classical James–Stein results are mostly developed within multivariate normal or approximately normal mean-estimation frameworks. For ease of exposition, we refer to our proposed shrinkage baseline as JS baseline as well, in order to highlight its conceptual similarity, though the resulting shrinkage rule is derived specifically for the RLVR setting and differs from this classical James–Stein formula.

\section{Conclusion}

This paper identifies an underexploited opportunity to reduce policy-gradient variance in critic-free RLVR and introduces a statistically principled, hyperparameter-free estimator that fully leverages the batch structure. The resulting shrinkage policy gradient estimator provides provable variance reduction while remaining a simple, drop-in replacement for existing RL pipelines. Theoretical derivations prove that the estimator has lower expectations of mean squared error compared with non-shrinkage counterparts, provably leading to lower policy gradient variance. Empirical results validate that the estimator consistently enhances RLVR training performance under different models, tasks and number of rollouts. Ablation analysis shows that the shrinkage baseline has broad applicability in reducing the variance of policy gradient estimator regardless of task difficulties or semantic heterogeneity. We expect the insights behind this approach to inspire further improvements in critic-free reinforcement learning.

\section*{Impact Statement}

This paper presents work whose goal is to advance the field of machine learning and large language models. There are many potential societal consequences of our work, none of which we feel must be specifically highlighted here.

\section*{Acknowledgement}

This work was partially supported by the National Science Foundation under Grants CCF-2106778 and DMS-2134080. This work used DeltaAI at NCSA through allocation CIS250428: Online Curriculum Learning for Large Language Models Reasoning and CIS250567: Training Reasoning Models for Best-of-N Performance from the Advanced Cyberinfrastructure Coordination Ecosystem: Services \& Support
(ACCESS) program (Boerner et al., 2023), which is supported by U.S. National Science Foundation
grants \#2138259, \#2138286, \#2138307, \#2137603, and \#2138296.

\bibliography{example_paper}
\bibliographystyle{icml2026}

\newpage
\appendix
\onecolumn

\clearpage
\newpage

\section{Theoretical Proof}
\label{sec:appendix_proof}
In this section, we use the notation $\vec Y - y_i^j$ to denote the set of all $y_{i'}^{j'}$ in which $(i',j')\neq (i,j)$.

\subsection{Proof of \Cref{prop:unbiased}}
\label{proof:unbiased}
Consider gradient update on each sample $\E_{\vec x, \vec Y}[(r_i^j-b_i^j)\nabla_\theta \log \pi_\theta(y_i^j|x_i)]$. We have 
\begin{align*}
    & \textstyle \E_{\vec x, \vec Y}[(r_i^j-b_i^j)\nabla_\theta \log \pi_\theta(y_i^j|x_i)] \\
    = & \textstyle \E_{\vec x, \vec Y}[r_i^j\nabla_\theta \log \pi_\theta(y_i^j|x_i)] - \E_{\vec x, \vec Y}[b_i^j\nabla_\theta \log \pi_\theta(y_i^j|x_i)] \\
    = & \textstyle \nabla_\theta J(\theta) - \E_{\vec x, \vec Y-y_i^j}\E_{y_i^j\sim \pi_\theta(\cdot\mid x_i)}[b_i^j\nabla_\theta \log \pi_\theta(y_i^j|x_i)] \\
    = & \textstyle \nabla_\theta J(\theta) - \E_{\vec x, \vec Y-y_i^j}\Bp{b_i^j\E_{y_i^j\sim \pi_\theta(\cdot\mid x_i)}[\nabla_\theta \log \pi_\theta(y_i^j|x_i)]} \\
    = & \textstyle \nabla_\theta J(\theta) - \E_{\vec x, \vec Y-y_i^j}\Bp{b_i^j\sum_{y_i^j}[\nabla_\theta\pi_\theta(y_i^j|x_i)]} \\
    = & \textstyle \nabla_\theta J(\theta).
\end{align*}
Since this holds for all $i,j$, we have
\begin{align*}
\textstyle \E_{\vec x, \vec Y}[g(\vec x, \vec Y;\theta)]
& = \textstyle \frac{1}{n}\sum_{i=1}^n \frac{1}{m} \sum_{j=1}^m \E_{\vec x, \vec Y}[(r_i^j-b_i^j)\nabla_\theta \log \pi_\theta(y_i^j|x_i)] \\
& = \textstyle \nabla_\theta J(\theta).
\end{align*}

\subsection{Proof of \Cref{lem:mse_motiv}}
\label{proof:mse_motiv}
Note that $b_i^{j,\mathrm{JS1}}=b_i^{1,\mathrm{JS1}}$ for all $i,j$. We can also rewrite the interpolation as 
\begin{align*}
    \textstyle b_i^{1,\mathrm{JS1}} = \Sp{1-\frac{n-1}{n}\lambda} \hat \mu_i + \frac{n-1}{n}\lambda \hat {\bar\mu}_{-i}. 
\end{align*}
We can let $\gamma=\frac{n-1}{n}\lambda$.
So we can rewrite the objective as
\begin{align*}
    \textstyle \mathrm{MSE}^{\mathrm{relax}} = &\textstyle \E_{\vec Y}\Mp{\frac{1}{n}\sum_{i=1}^n \Sp{\mu_i-b_i^{1,\mathrm{JS1}}}^2} \\
    = & \textstyle \frac{1}{n}\sum_{i=1}^n \E_{\vec Y}\Mp{((1-\gamma)(\mu_i-\hat \mu_i)+ \gamma (\mu_i-\hat{\bar \mu}_{-i}))^2} \\
    = & \textstyle \frac{1}{n}\sum_{i=1}^n \Bp{(1-\gamma)^2\E_{\vec Y}\Mp{(\mu_i-\hat \mu_i)^2} +2\gamma(1-\gamma)\E_{\vec Y}\Mp{(\mu_i-\hat \mu_i)(\mu_i-\hat{\bar \mu}_{-i})} + \gamma^2 \E_{\vec Y}\Mp{(\mu_i-\hat{\bar \mu}_{-i})^2} } \\
    = & \textstyle \frac{1}{n}\sum_{i=1}^n \Bp{(1-\gamma)^2\var_{\vec Y}[\hat \mu_i] +2\gamma(1-\gamma)\E_{\vec Y}\Mp{\mu_i-\hat \mu_i}\E_{\vec Y}\Mp{\mu_i-\hat{\bar \mu}_{-i}} + \gamma^2 \E_{\vec Y}\Mp{(\mu_i-\hat{\bar \mu}_{-i})^2} } \\
    = & \textstyle \frac{1}{n}\sum_{i=1}^n \Bp{(1-\gamma)^2\frac{1}{m}\sigma_i^2  + \gamma^2 \E_{\vec Y}\Mp{(\mu_i-\hat{\bar \mu}_{-i})^2} }.
\end{align*}

Let $\bar \mu_{-i}:= \frac{1}{n-1}\sum_{i'\neq i}\mu_{i'}$. For the second term in the summation, note that
\begin{align*}
\textstyle \E_{\vec Y}\Mp{(\mu_i-\hat{\bar \mu}_{-i})^2} 
& = \textstyle \E_{\vec Y}\Mp{((\mu_i-\bar \mu_{-i}) + (\bar \mu_{-i} - \hat{\bar \mu}_{-i}))^2} \\
& = \textstyle (\mu_i-\bar \mu_{-i})^2 + \E_{\vec Y}\Mp{(\bar \mu_{-i} - \hat{\bar \mu}_{-i})^2} \\
& = \textstyle \Sp{\frac{n}{n-1}}^2 (\mu_i-\bar \mu)^2 + \var_{\vec Y}\Mp{\hat{\bar \mu}_{-i}} \\
& = \textstyle \Sp{\frac{n}{n-1}}^2 (\mu_i-\bar r)^2 + \frac{1}{(n-1)^2}\sum_{i'\neq i}\var_{\vec Y}\Mp{\hat \mu_i}.
\end{align*}
Therefore, we have
\begin{align*}
&\textstyle \E\Mp{\frac{1}{n}\sum_{i=1}^n \Sp{\mu_i-b_i^{1, \mathrm{JS1}}}^2} \\
 = & \textstyle \frac{1}{n}\sum_{i=1}^n \Bp{(1-\gamma)^2\frac{1}{m}\sigma_i^2  + \gamma^2 \Mp{\Sp{\frac{n}{n-1}}^2 (\mu_i-\bar \mu)^2 + \frac{1}{(n-1)^2}\sum_{i'\neq i}\frac{1}{m}\sigma_i^2}} \\
 = & \textstyle \frac{1}{n}(1-\gamma)^2\sum_{i=1}^n\frac{1}{m}\sigma_i^2 + \frac{n}{(n-1)^2} \gamma^2 \sum_{i=1}^n (\mu_i-\bar \mu)^2 + \frac{1}{n(n-1)^2}\gamma^2 \sum_{i=1}^n\sum_{i'\neq i}\frac{1}{m}\sigma_i^2 \\
 = & \textstyle \frac{1}{n}(1-\gamma)^2\sum_{i=1}^n\frac{1}{m}\sigma_i^2 + \frac{n}{(n-1)^2} \gamma^2 \sum_{i=1}^n (\mu_i-\bar \mu)^2 + \frac{1}{n(n-1)}\gamma^2 \sum_{i=1}^n\frac{1}{m}\sigma_i^2 \\
 = & \textstyle \frac{n}{n-1}(s+v)\gamma^2 -2v \gamma + v.
\end{align*}
This is a quadratic function of $\gamma$, and so the global minimum is $\gamma^*:=\frac{n-1}{n}\frac{v}{s+v}$.

\subsection{Proof of \Cref{lem:mse_unbias}}
\label{proof:mse_unbias}
Recall \Cref{eq:baseline_unbias}:
\begin{align*}
    \textstyle b_i^{j,\mathrm{JS2}}=(1-\lambda_i^{j})\hat \mu_i^{-j}+\lambda_i^j \hat{\bar \mu}_{-i}.
\end{align*} 
Since each baseline $b_i^{j,\mathrm{JS2}}$ has its own shrinkage coefficient $\lambda_i^j$, we only need to minimize each square error term $\E_{\vec x, \vec Y }[\mu_i- b_i^{j,\mathrm{JS2}}]$ in order to minimize the whole objective $\mathrm{MSE}$.

Then for any $i,j$, we have 
\begin{align*}
&\textstyle \E_{\vec x, \vec Y }[(\mu_i-b_i^j)^2] \\
=& \textstyle \E_{\vec x, \vec Y }[(1-\lambda_i^j)(\mu_i-\hat \mu_i^{-j})+\lambda_i^j (\mu_i-\hat{\bar \mu}_{-i}))^2] \\
=& \textstyle \E_{\vec x}\E_{\vec Y }\Mp{(1-\lambda_i^j)^2(\mu_i-\hat \mu_i^{-j})^2+(\lambda_i^j)^2 (\mu_i-\hat{\bar \mu}_{-i})^2+2\lambda_i^j(1-\lambda_i^j)(\mu_i-\hat \mu_i^{-j})(\mu_i-\hat{\bar \mu}_{-i})} \\
= & \textstyle \E_{\vec X}\Bp{(1-\lambda_i^j)^2\E_{\vec Y}\Mp{(\mu_i-\hat \mu_i^{-j})^2}+(\lambda_i^j)^2\E_{\vec Y}\Mp{(\mu_i-\hat{\bar \mu}_{-i})^2} + 2\lambda_i^j(1-\lambda_i^j)\E_{\vec Y}\Mp{\mu_i-\hat \mu_i^{-j}}\E_{\vec Y}\Mp{\mu_i-\hat{\bar \mu}_{-i}} } \\
= & \textstyle \E_{\vec X}\Bp{(1-\lambda_i^j)^2\E_{\vec Y}\Mp{(\mu_i-\hat \mu_i^{-j})^2}+(\lambda_i^j)^2\E_{\vec Y}\Mp{(\mu_i-\hat{\bar \mu}_{-i})^2}}.
\end{align*}
For the first term in the summation, we have
\begin{align*}
\textstyle \E_{\vec Y}\Mp{(\mu_i-\hat \mu_i^{-j})^2} = \var_{\vec Y}[\hat \mu_i^{-j}] = \frac{1}{m-1}\sigma^2(x_i).
\end{align*}
Recall that $\sigma^2(x_i) = \var_{y\sim \pi_\theta(\cdot\mid x_i)}[r(x_i,y)]$.
For the second term, we have
\begin{align*}
\textstyle \E_{\vec Y}\Mp{(\mu_i-\hat{\bar \mu}_{-i})^2}
& = \textstyle \E_{\vec Y}\Mp{((\mu_i-\bar \mu_{-i})+(\bar\mu_{-i}-\hat{\bar \mu}_{-i}))^2} \\
& = \textstyle (\mu_i-\bar \mu_{-i})^2+ \var_{\vec Y}[\hat{\bar \mu}_{-i}] \\
& = \textstyle (\mu_i-\bar \mu_{-i})^2 + \frac{1}{(n-1)^2}\sum_{i'\neq i}\var_{\vec Y}[\hat \mu_{i'}] \\
& = \textstyle (\mu_i-\bar \mu_{-i})^2 + \frac{1}{(n-1)^2}\sum_{i'\neq i}\frac{1}{m-1}\sigma^2(x_{i'})
\end{align*}
Combining together, we have
\begin{align*}
&\textstyle \E_{\vec x, \vec Y }[(\mu_i-b_i^j)^2] \\
=&\textstyle (1-\lambda_i^j)^2\E_{\vec x}\Mp{\frac{1}{m-1}\sigma^2(x_i) }+(\lambda_i^j)^2\E_{\vec x}\Mp{ (\mu_i-\bar \mu_{-i})^2 + \frac{1}{(n-1)^2}\sum_{i'\neq i}\frac{1}{m-1}\sigma^2(x_{i'}) } \\
= &\textstyle \frac{1}{m-1}\E_{x\sim\cD}[\sigma^2(x)](1-\lambda_i^j)^2+\frac{1}{(n-1)(m-1)}\E_{x\sim\cD}[\sigma^2(x)](\lambda_i^j)^2 + \frac{n}{n-1}\var_{x\sim\cD}[\mu(x)](\lambda_i^j)^2 \\
= & \textstyle \frac{n}{n-1}(s_2+v_2)(\lambda_i^j)^2-2v_2\lambda_i^j+v_2.
\end{align*}
Therefore, optimal $\lambda_i^j$ is 
\begin{align*}
\textstyle (\lambda_i^j)^* = \frac{n-1}{n} \frac{v_2}{s_2+v_2}.
\end{align*}
This holds for all $i,j$, so we complete the proof.

\clearpage
\newpage

\section{Direct Verification of Variance Reduction}
\label{app:direct-verify}

\subsection{Reduction of Mean Squared Error with Ground-Truth Value Function}
\label{app:mse}
  
  \begin{figure}[htbp]
    \centering
    \includegraphics[width=0.5\textwidth]{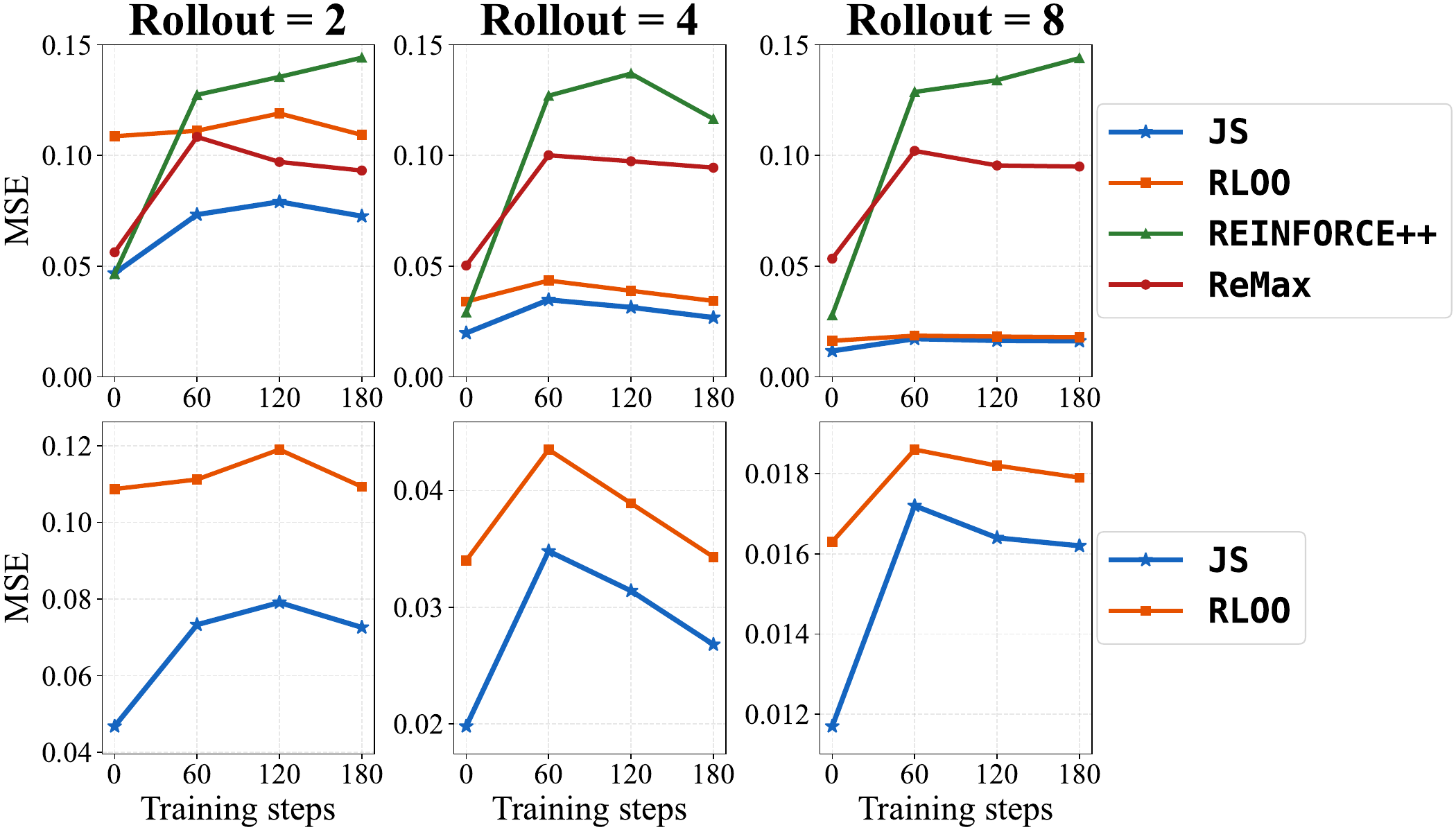}
    \caption{{MSE between value score and estimated baseline under different rollout budgets during training. 
    The results are based on the average of 20 randomly selected batches from DAPO17k, and weights from the experiments on Qwen3-4B-Base. 
    With baseline shrinkage, the estimation is consistently closer to value score. 
    For 2 rollouts, 4 rollouts, 8 rollouts, mean squared error for shrinkage baseline estimator are 39.4\%, 25.1\% and 13.4\% lower than RLOO estimator, respectively.}}
    \label{fig:mse_estimation}
  \end{figure}
    
    Moreover, we estimate the value function of each question be monte-carlo sampling. For each question, we generate 128 trajectories and compute the average reward score as an accurate estimation. After that, we sample another batch of responses under fixed smaller numbers of rollouts, and computed the mean squared error between different baselines and estimated value in a batch of 64 questions, same as the training setting. The MSE metric for RLOO, REINFORCE++, ReMax and JS baseline throughout training are shown as \Cref{fig:mse_estimation}. Under different rollout budgets, the JS baseline consistently shows smallest deviation with the value score compared with RLOO, REINFORCE++ and ReMax. These results support our theoretical derivations and highlight the advantage of the shrinkage baseline estimation for reinforcement finetuning under varying rollout counts.

\subsection{Reduction of Squared L2 Norm with Ground-Truth Policy Gradient}
\label{app:gt-pg}
    
    \begin{figure}[htbp]
      \centering
      \includegraphics[width=0.5\textwidth]{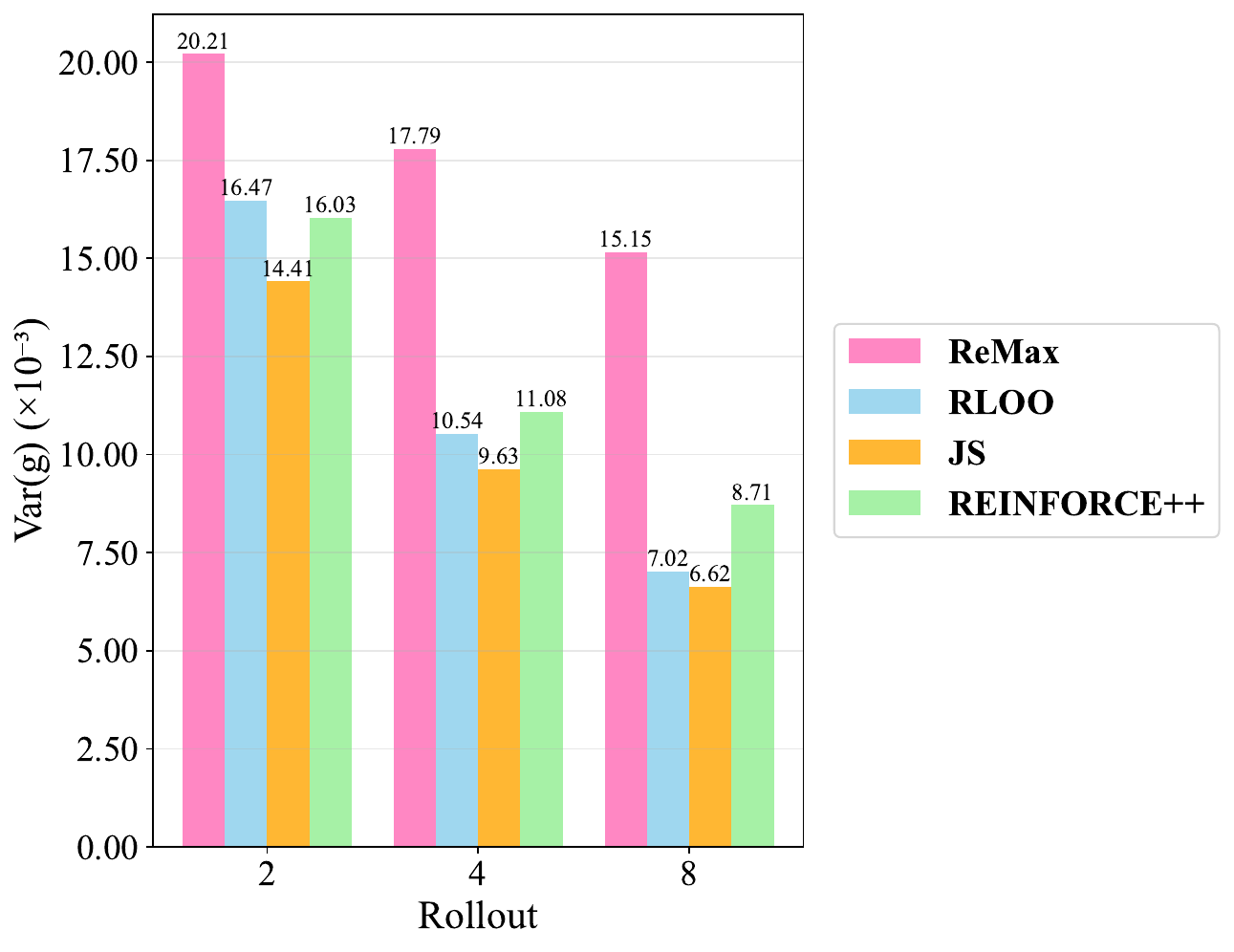}
      \caption{Comparison of different baselines on variance of policy gradient, obtained by computing squared L2 norm with ground truth policy gradients.}
      \label{fig:gt-pg}
    \end{figure}
    
    We further obtain a precise estimate of the ground-truth policy gradient $\nabla_\theta J(\theta)$ and directly compare the variance of the policy gradient under different baselines using \Cref{eqn:varG}. We first select 128 questions from the DAPO17k dataset and generate 256 responses for each question using the Qwen3-4B-Base model. For each question, by using all 256 rollouts to compute the policy gradient (\Cref{eq:grad}), we obtain a precise estimate of the ground-truth $\nabla_\theta J(\theta)$. We then randomly sample 64 questions and $n$ (much smaller than 256) responses for each of them from the large response dataset we created. A small batch of $64 \times n$ samples is formed, similar to RL training setups in practice. Based on this small batch, we compute the policy gradients $g_{\text{RLOO}}$, $g_{\text{REINFORCE++}}$, $g_{\text{ReMax}}$ and $g_{\text{JS}}$ using different baselines, and compute the variance of the policy gradient by taking the squared L2 norm between $g$ and $\nabla_\theta J(\theta)$ (\Cref{eqn:varG}). In \ref{fig:gt-pg}, we report the average squared L2 norm over 20 small batches with rollout numbers $n = 2, 4, 8$. We can see that the use of the shrinkage baseline reduces the L2 norm between the estimated policy gradient and the ground truth by 12.5\%, 8.6\%, and 5.7\%, respectively, providing direct evidence for improved training stability.

\clearpage
\newpage

\section{Ablation Study}
\label{app:ablation}

\subsection{Ablation Study on Dataset Difficulty Heterogeneity}
\label{app:data-heter}

\begin{figure}[htbp]
  \centering
  \includegraphics[width=0.5\textwidth]{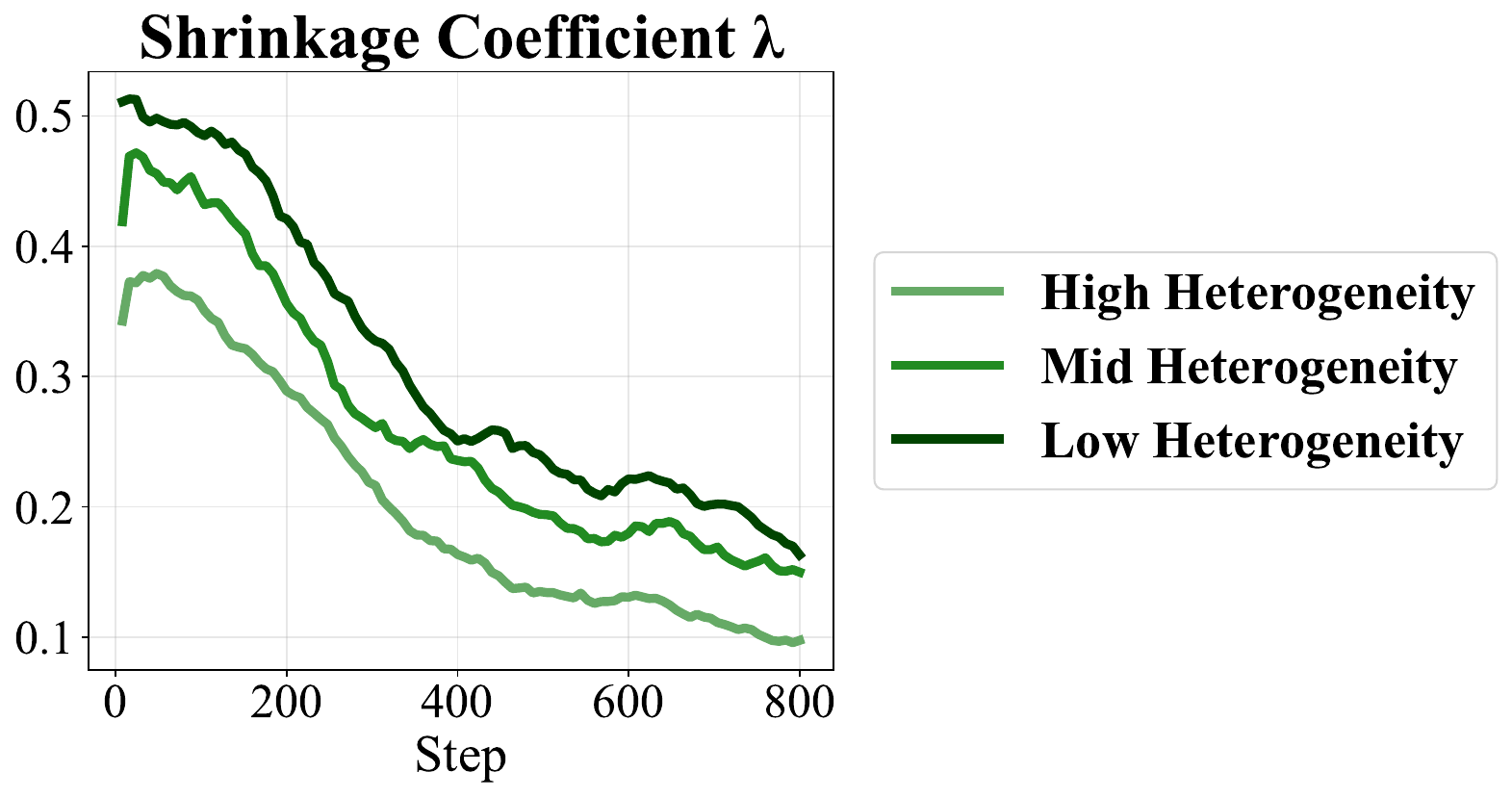}
  \caption{JS shrinkage coefficient during training in different difficulty heterogeneity.}
  \label{fig:data-heter}
\end{figure}

\begin{table}[htbp]
  \centering
  \small
  \begin{tabular}{lccc}
    \toprule
    \textbf{Method} & \textbf{Low} & \textbf{Mid} & \textbf{High} \\
    \midrule
    RLOO & 20.25 & 17.69 & 31.13 \\
    JS   & 25.69 & 23.85 & 34.25 \\
    \bottomrule
  \end{tabular}
  \caption{Average test set accuracy (\%) under different heterogeneity settings.}
  \label{tab:low-mid-high}
\end{table}

We conduct an ablation study on the effect of the JS baseline under varying levels of dataset heterogeneity. For the Knights-and-Knaves dataset, a larger problem size (i.e., the number of propositions given) indicates higher complexity and difficulty. Therefore, we consider three training settings by mixing data with different difficulty levels: low heterogeneity (KnK-4 \& KnK-5), medium heterogeneity (KnK-3 \& KnK-7), and high heterogeneity (KnK-2 \& KnK-9). For each dataset, we train a Qwen2.5-1.5B-Instruct model \citep{qwen2.5} for 800 gradient steps and evaluate it on 200 questions drawn from the same distribution as the training data. \Cref{fig:data-heter} shows the shrinkage coefficient for the different datasets, and \Cref{tab:low-mid-high} reports the test accuracy in the different settings. We observe that higher heterogeneity in the difficulty distribution leads to an adaptively smaller shrinkage, i.e., the algorithm behaves closer to vanilla RLOO. This adaptive adjustment mechanism yields an approximately optimal baseline across different training data distributions, resulting in consistent performance gains.

\subsection{Ablation Study on Semantic Heterogeneity}
\label{app:semantic-heter}

\begin{table}[htbp]
  \centering
  \small
  \setlength{\tabcolsep}{6pt}
  \renewcommand{\arraystretch}{1.05}
  \begin{tabular}{@{}lcc@{}}
    \toprule
    Dataset & Pass Rate (\%) & $\lambda$ \\
    \midrule
    KnK-4       & 13.35 & 0.43 \\
    Countdown-4 & 14.62 & 0.41 \\
    \bottomrule
  \end{tabular}
  \caption{Qwen2.5-7B-Instruct pass rate (\%) and shrinkage coefficient ($\lambda$) on two datasets.}
  \label{tab:passrate-jscoef-knk4-countdown3}
\end{table}

\begin{figure}[htbp]
  \centering
  \includegraphics[width=0.4\textwidth]{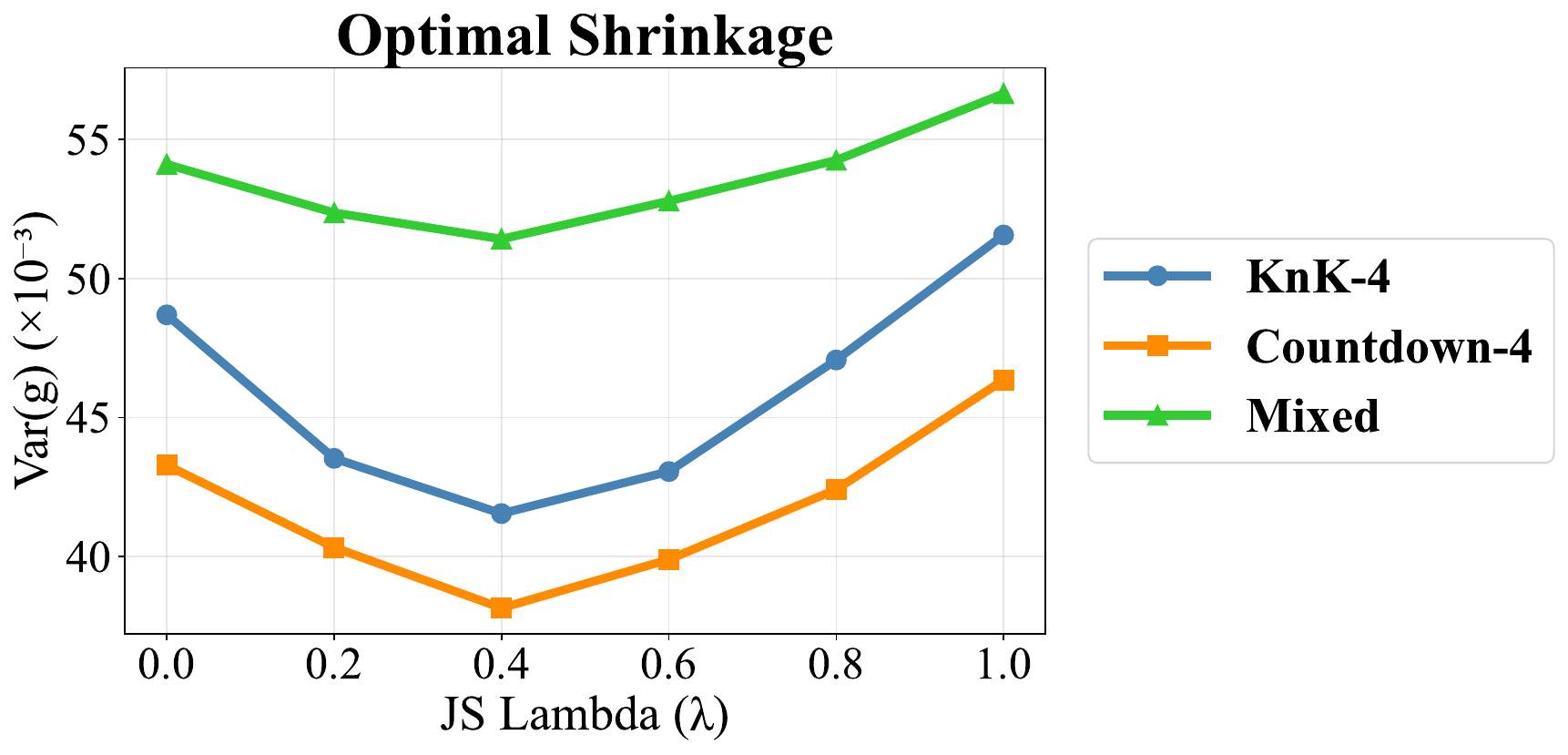}
  \caption{Effect of semantic heterogeneity on policy-gradient variance reduction with shrinkage baselines.}
  \label{fig:optimal_shrinkage}
\end{figure}

In addition, we analyze how semantic heterogeneity of prompts affects variance reduction when using shrinkage baselines. We select three training datasets from our earlier experiments: KnK-4, Countdown-4, and a mixed dataset formed by combining 50\% KnK-4 and 50\% Countdown-4 problems. We run all experiments on Qwen2.5-7B-Instruct. By design, KnK-4 and Countdown-4 have similar pass rates and similar shrinkage coefficients on this model (\Cref{tab:passrate-jscoef-knk4-countdown3}). Therefore, mixing the two tasks \textbf{increases prompt semantic heterogeneity} while keeping the \textbf{reward distribution and shrinkage coefficient approximately matched}. Following the procedure in Appendix~\ref{app:gt-pg}, we estimate the variance of the policy-gradient estimator on Qwen2.5-7B-Instruct under different shrinkage coefficients $\lambda$ for all three datasets, with a fixed rollout number of 4. As shown in Figure~\ref{fig:optimal_shrinkage}, relative to the RLOO baseline ($\lambda=0.0$), JS shrinkage with $\lambda=0.4$ reduces policy-gradient variance by 14.7\% on KnK-4 and 11.8\% on Countdown-4. On the mixed dataset, JS shrinkage still reduces variance, but the improvement is smaller (5.32\%). However, $\lambda \approx 0.4$ remains the variance-minimizing choice in all three settings. These results suggest that semantic heterogeneity mainly influences the magnitude of variance reduction, while a James-Stein-style interpolation between the prompt-level mean and the batch-level mean continues to minimize gradient variance, regardless of how semantically heterogeneous the training data is.

\clearpage
\newpage

\section{Experimental Details}
\label{sec:appendix_exp}

\subsection{Code Implementation}

Below is the core python implementation of computing the advantage according to the shrinkage baseline. It only requires several lines of code and negligible additional computation.

\begin{lstlisting}[caption={Shrinkage Baseline Advantage Estimator}, label={lst:js-adv}]
prompt_mean = torch.mean(rewards, dim=1)  # [n, m] => [n]
prompt_var = torch.var(rewards, dim=1, unbiased=True) / m

# Compute LOO batch mean and JS lambda
loo_means, js_lambdas = [], []
for i in range(n):
    other = torch.cat([prompt_mean[:i], prompt_mean[i + 1:]])
    batch_loo_mean = torch.mean(other)
    v_square_i = torch.mean(torch.cat([prompt_var[:i], prompt_var[i + 1:]]))
    s_square_i = torch.mean((other - batch_loo_mean) ** 2)
    js_lambda_i = v_square_i / (v_square_i + s_square_i)
    js_lambda_i *= (n - 1) / n
    loo_means.append(batch_loo_mean)
    js_lambdas.append(js_lambda_i)

loo_means = torch.stack(loo_means)    # [n]
js_lambdas = torch.stack(js_lambdas)  # [n]
rloo_baseline = (torch.sum(rewards, dim=1, keepdim=True) - rewards) / (m - 1)
js_baseline = rloo_baseline * (1 - js_lambdas[:, None]) + loo_means[:, None] * js_lambdas[:, None]
advantage = rewards - js_baseline
\end{lstlisting}

\subsection{Variance Recording}
\label{app:variance-recording}

To track the gradient variance during training, we derive an unbiased estimator based on micro-batch gradients. Let i.i.d. micro-batch gradients $g_i\in\mathbb{R}^P$ for $i=1,\dots,m$, with $\mu=\E[g_i]$ and $\Sigma=\Cov(g_i)$. Let $g=\frac{1}{m}\sum_{i=1}^m g_i$ denote the batch gradient, and $\bar g=\frac{1}{m}\sum_{i=1}^m g_i$ the sample mean.

We first express the target quantity:
\begin{align*}
\textstyle \Var(g)
= \tr(\Cov(g))
= \tr(\Cov(\tfrac{1}{m}\sum_{i=1}^m g_i))
= \tr(\tfrac{1}{m^2}\sum_{i=1}^m \Cov(g_i))
= \frac{1}{m}\tr(\Sigma).
\end{align*}

Using the identity $\sum_{i=1}^m \|g_i-\bar g\|^2=\sum_{i=1}^m \|g_i\|^2 - m\|\bar g\|^2$, we take expectations term by term:
\begin{align*}
\textstyle \E[\sum_{i=1}^m \|g_i-\bar g\|^2]
= \sum_{i=1}^m \E\|g_i\|^2 - m\,\E\|\bar g\|^2.
\end{align*}

For the two moments, we have
\begin{align*}
\textstyle \E\|g_i\|^2 = \tr(\Sigma)+\|\mu\|^2, \qquad
\E\|\bar g\|^2
= \E\|\frac{1}{m}\sum_{i=1}^m g_i\|^2
= \frac{1}{m}\tr(\Sigma)+\|\mu\|^2.
\end{align*}

Substituting back gives
\begin{align*}
\textstyle \E[\sum_{i=1}^m \|g_i-\bar g\|^2]
= m(\tr(\Sigma)+\|\mu\|^2) - m(\frac{1}{m}\tr(\Sigma)+\|\mu\|^2)
= (m-1)\tr(\Sigma).
\end{align*}

Dividing by $(m-1)$ yields the unbiased sample-trace of $\Sigma$:
\begin{align*}
\textstyle \E[\frac{1}{m-1}\sum_{i=1}^m \|g_i-\bar g\|^2] = \tr(\Sigma).
\end{align*}

Since $\Var(g)=\frac{1}{m}\tr(\Sigma)$, multiplying by $\frac{1}{m}$ produces the desired unbiased estimator of $\Var(g)$:
\begin{align*}
\textstyle \E[\frac{1}{m}\cdot\frac{1}{m-1}\sum_{i=1}^m \|g_i-\bar g\|^2]
= \frac{1}{m}\tr(\Sigma) = \Var(g).
\end{align*}

Finally, using $\sum_{i=1}^m \|g_i-\bar g\|^2=\sum_{i=1}^m \|g_i\|^2-\frac{1}{m}\|\sum_{i=1}^m g_i\|^2$, the estimator can be written in the form of:
\begin{align*}
\textstyle \widehat{\Var(g)}
= \frac{1}{m}\cdot\frac{1}{m-1}(\sum_{i=1}^m \|g_i\|^2
- \frac{1}{m}\|\sum_{i=1}^m g_i\|^2).
\end{align*}

Below is the pseudo code of gradient variance estimation:

\begin{lstlisting}[caption={Gradient Variance Estimation}]
sum_sq = 0.0                 # accumulates sum_i ||g_i||^2
sum_g = zeros_like(vector)   # accumulates sum_i g_i  (shape: (P,))

for i in range(1, m + 1):
    g_i = compute_flattened_gradient_for_microbatch(i)  # shape: (P,)
    sum_sq += dot(g_i, g_i)  # scalar: ||g_i||^2
    sum_g  += g_i            # vector: sum of g_i

# Sample-trace of covariance at micro-batch level
# (1/(m-1)) * sum_i ||g_i - g_bar||^2
trace_cov_micro = (sum_sq - dot(sum_g, sum_g) / m) / (m - 1)

# Unbiased estimate of batch-gradient variance trace
# Var(g) = tr(Cov(g)), because Cov(g) = (1/m) * Cov(g_i)
var_g_trace_estimate = (1.0 / m) * trace_cov_micro
\end{lstlisting}

\clearpage

\subsection{Supplementary Results}
\label{app:supplementary-results}

We aggregate additional experimental results in this section.

\subsubsection{Comparison with Different Baselines}

We train Qwen2.5-0.5B-Instruct model on GSM8k dataset for 500 steps. GSM8k consists of 7,473 questions in the training set and 1,319 in test set. For each rollout batch, we sample 64 distinct prompts, and for each prompt we generate $m \in \{2, 4, 8\}$ responses with official template of GSM8k. Each experiment is repeated across five random seeds \{0,1,2,3,4\}. Summary of hyperparameters and configurations is provided in \Cref{tab:exp-config}. Detailed numbers are in \Cref{tab:appendix-data}.

\begin{table}[h]
  \centering
  \tiny
  \setlength{\tabcolsep}{2.5pt}
  \renewcommand{\arraystretch}{1.2}
  \caption{Test Accuracy (\%) across 5 runs for different algorithms under 2, 4, and 8 Generations. Each cell shows 5 accuracy values (\%) at steps 100 to 500. Initial test accuracy is 40.03\% for all.} 
  \label{tab:appendix-data}
  \begin{tabular}{l|p{2.9cm}|p{2.9cm}|p{2.9cm}}
    \toprule
    \textbf{Baseline} & \textbf{2 Generations} & \textbf{4 Generations} & \textbf{8 Generations} \\
    \midrule
    JS &
    48.29, 50.96, 53.56, 54.45, 54.68 \newline
    47.55, 52.36, 53.77, 54.56, 56.12 \newline
    50.11, 51.84, 52.06, 54.18, 55.27 \newline
    49.10, 51.92, 54.12, 55.09, 54.97 \newline
    48.58, 52.01, 53.66, 54.03, 55.07 &
    50.22, 52.53, 56.56, 57.29, 58.03 \newline
    50.51, 54.48, 56.56, 56.57, 57.80 \newline
    50.83, 53.45, 55.41, 55.95, 56.92 \newline
    51.43, 54.63, 55.85, 56.39, 56.29 \newline
    52.37, 55.15, 56.59, 56.48, 57.59 &
    52.19, 56.77, 57.65, 59.42, 60.12 \newline
    53.28, 56.15, 56.53, 57.42, 57.89 \newline
    53.39, 55.19, 56.92, 58.67, 58.73 \newline
    52.77, 55.31, 58.13, 58.04, 58.24 \newline
    52.51, 55.03, 57.03, 58.23, 59.68 \\
    \midrule
    BLOO &
    49.42, 50.80, 51.88, 52.23, 54.61 \newline
    48.88, 51.43, 52.32, 53.39, 53.39 \newline
    48.22, 50.42, 52.99, 54.41, 54.30 \newline
    48.08, 49.95, 53.15, 53.43, 53.06 \newline
    48.57, 51.46, 52.08, 53.98, 55.62 &
    49.32, 52.65, 54.19, 55.16, 56.13 \newline
    50.46, 53.39, 55.85, 55.11, 55.90 \newline
    49.64, 52.43, 54.36, 54.47, 55.95 \newline
    50.28, 52.86, 54.53, 54.51, 55.98 \newline
    49.55, 52.53, 54.71, 54.91, 56.35 &
    52.10, 54.61, 55.91, 56.91, 57.54 \newline
    51.68, 54.59, 56.63, 56.07, 57.32 \newline
    50.99, 54.25, 55.13, 55.68, 57.02 \newline
    51.33, 54.71, 56.15, 55.16, 57.47 \newline
    51.08, 54.98, 56.06, 56.95, 56.74 \\
    \midrule
    RLOO &
    49.23, 51.13, 52.02, 52.54, 54.89 \newline
    48.48, 51.95, 52.84, 53.33, 53.56 \newline
    47.95, 50.99, 53.37, 53.99, 54.63 \newline
    47.98, 50.75, 51.81, 54.30, 54.91 \newline
    47.46, 51.21, 51.64, 52.86, 54.45 &
    51.01, 53.12, 55.62, 56.26, 56.10 \newline
    50.84, 53.18, 54.24, 56.10, 55.28 \newline
    50.42, 53.39, 54.53, 55.47, 56.60 \newline
    50.10, 53.60, 54.59, 56.16, 56.92 \newline
    51.19, 54.36, 56.29, 56.16, 56.80 &
    52.99, 55.31, 56.89, 58.18, 58.50 \newline
    52.92, 56.71, 56.71, 57.24, 58.01 \newline
    52.25, 54.54, 56.29, 57.20, 58.18 \newline
    52.62, 55.10, 56.03, 57.88, 58.63 \newline
    52.34, 55.38, 56.86, 56.85, 58.21 \\
    \midrule
    GRPO &
    48.67, 51.99, 52.99, 54.72, 54.41 \newline
    47.73, 52.42, 53.15, 54.06, 52.87 \newline
    48.92, 51.08, 54.24, 54.12, 54.34 \newline
    48.57, 51.72, 50.70, 52.89, 53.01 \newline
    48.77, 51.52, 53.24, 53.49, 53.79 &
    51.39, 53.54, 55.56, 55.94, 55.54 \newline
    51.31, 52.96, 55.15, 55.79, 57.39 \newline
    49.75, 53.51, 54.38, 54.57, 55.21 \newline
    50.05, 54.43, 56.62, 57.77, 56.88 \newline
    51.63, 53.04, 54.47, 55.09, 56.19 &
    51.96, 56.07, 57.10, 58.27, 58.20 \newline
    52.52, 55.53, 56.51, 57.80, 59.06 \newline
    52.43, 55.57, 57.12, 58.37, 59.05 \newline
    52.43, 55.77, 57.56, 57.54, 57.21 \newline
    52.48, 55.31, 57.83, 57.85, 57.91 \\
    \midrule
    ReMax &
    49.16, 52.12, 52.40, 52.54, 54.83 \newline
    48.76, 52.68, 53.37, 54.39, 54.10 \newline
    48.79, 52.10, 52.45, 53.75, 55.65 \newline
    48.57, 51.52, 52.98, 52.40, 54.06 \newline
    48.92, 51.63, 52.54, 53.33, 54.86 &
    50.05, 53.37, 55.35, 55.95, 56.33 \newline
    49.08, 51.15, 53.92, 55.69, 55.94 \newline
    50.19, 52.83, 54.98, 54.59, 57.12 \newline
    48.78, 51.31, 54.12, 54.98, 56.57 \newline
    49.48, 53.80, 54.16, 55.25, 56.40 &
    51.04, 54.71, 56.65, 57.10, 57.81 \newline
    52.01, 55.71, 56.29, 57.10, 59.38 \newline
    51.55, 54.22, 55.66, 57.26, 57.62 \newline
    51.05, 54.39, 55.69, 55.36, 56.19 \newline
    51.32, 54.84, 55.99, 56.87, 57.81 \\
    \midrule
    REINFORCE++ &
    48.86, 51.90, 53.80, 53.57, 55.01 \newline
    49.61, 52.18, 53.06, 54.86, 55.27 \newline
    48.19, 50.54, 52.66, 53.74, 53.51 \newline
    49.51, 51.96, 52.27, 54.63, 55.33 \newline
    48.76, 51.35, 52.46, 53.98, 54.96 &
    50.05, 52.87, 53.77, 55.44, 56.79 \newline
    49.95, 53.25, 54.98, 54.97, 53.69 \newline
    49.11, 53.33, 55.19, 55.72, 56.04 \newline
    49.63, 52.24, 53.72, 55.13, 54.01 \newline
    49.52, 52.68, 54.86, 55.21, 55.82 &
    51.51, 54.50, 55.59, 58.10, 57.92 \newline
    51.17, 54.57, 57.22, 57.15, 57.24 \newline
    52.08, 55.82, 57.16, 57.13, 57.00 \newline
    51.95, 54.84, 57.03, 57.77, 57.04 \newline
    50.99, 53.82, 56.94, 57.23, 57.32 \\
    \bottomrule
  \end{tabular}
\end{table}

\subsubsection{Knights-and-Knaves Subset Results}

\begin{table}[h]
  \centering
  \setlength{\tabcolsep}{6pt} 
  \renewcommand{\arraystretch}{1.05} 

  \captionsetup{width=.95\linewidth,skip=4pt,justification=centering}
  \caption{Detailed test accuracy (\%) on different subsets for Knights-and-Knaves experiments. Larger number in KnK means the task is more complex.}
  \label{tab:knk-train-easy-hard}

  \begin{tabular}{@{}l ccc ccc@{}}
    \toprule
    \multirow{2}{*}{Algorithm}
      & \multicolumn{3}{c}{\textbf{Train on KnK-Easy}} 
      & \multicolumn{3}{c}{\textbf{Train on KnK-Hard}} \\
    \cmidrule(lr){2-4} \cmidrule(lr){5-7}
      & KnK-4 & KnK-5 & KnK-6
      & KnK-6 & KnK-7 & KnK-8 \\
    \midrule
    RLOO & 58.38 & 49.89 & 41.25 & 39.50 & 30.00 & 16.75 \\
    JS & 75.87 & 66.16 & 55.25 & 42.13 & 32.50 & 25.38 \\
    \bottomrule
  \end{tabular}
\end{table}

\clearpage
\subsubsection{Additional Training Curves}

\begin{figure}[htbp]
  \centering
  \includegraphics[width=0.6\textwidth]{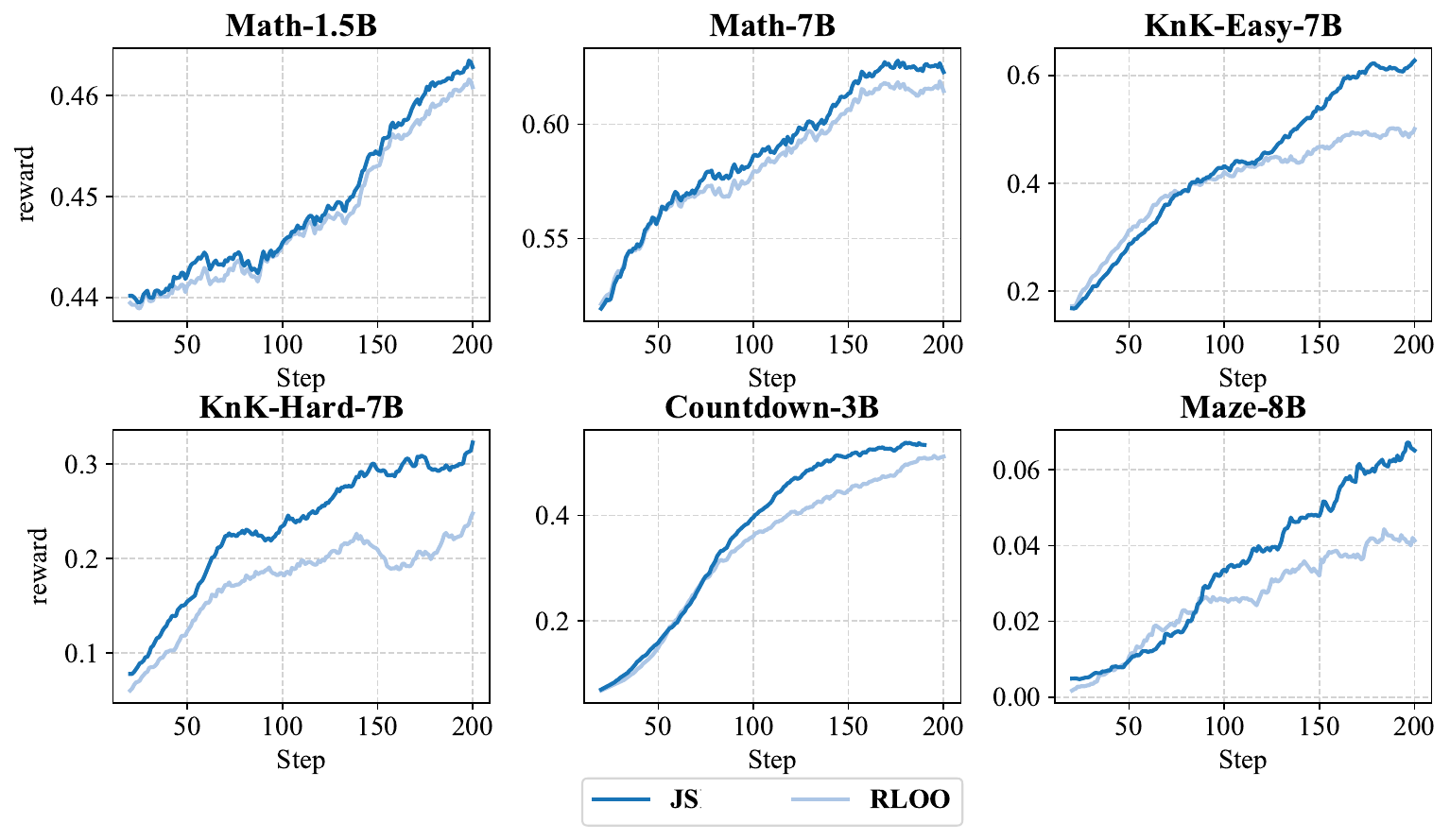}
  \caption{Moving average reward curves during training, compared with shrinkage baseline and RLOO baseline.}
  \label{fig:reward_curves}
\end{figure}

\begin{figure}[htbp]
  \centering
  \includegraphics[width=0.5\textwidth]{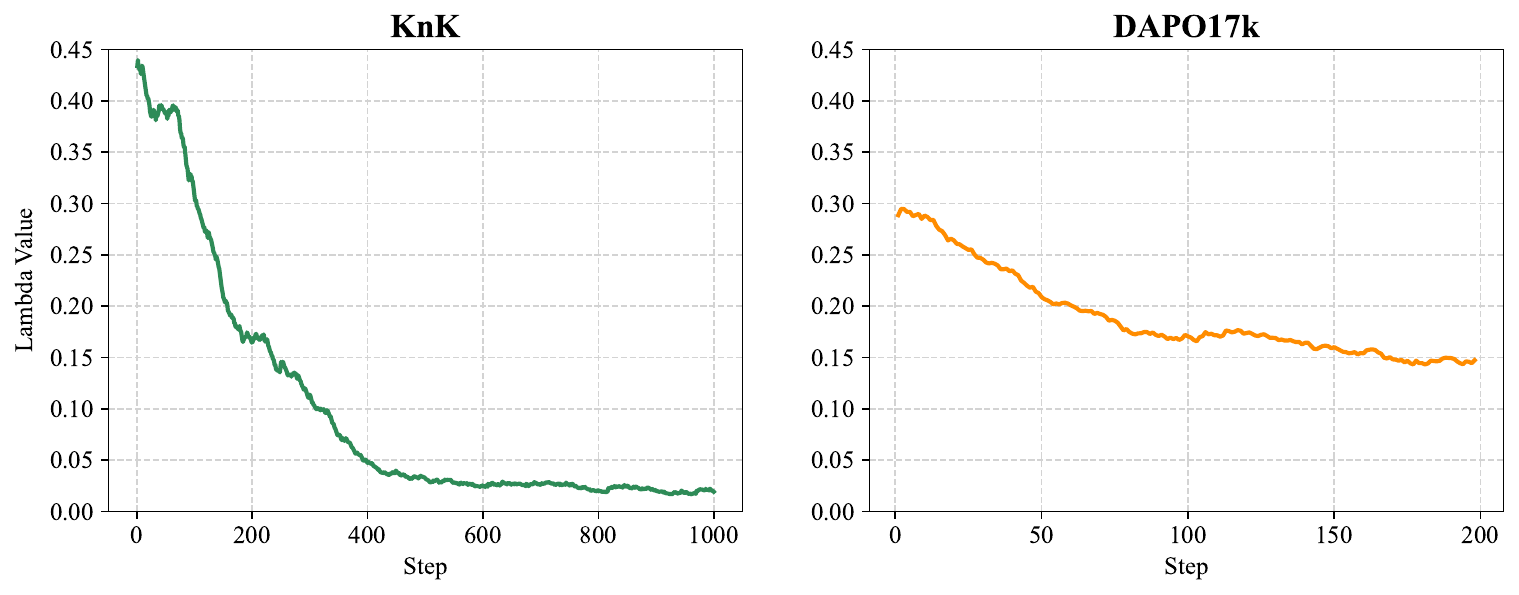}
  \caption{The moving average shrinkage coefficient curves on different datasets. Left: KnK dataset; Right: DAPO17k dataset. Adaptive shrinkage enables dynamically optimal baseline estimation that adjusts to the data distributions and training progresses.}
  \label{fig:more_lambda_curves}
\end{figure}

\begin{figure}[htbp]
  \centering
  \includegraphics[width=0.6\textwidth]{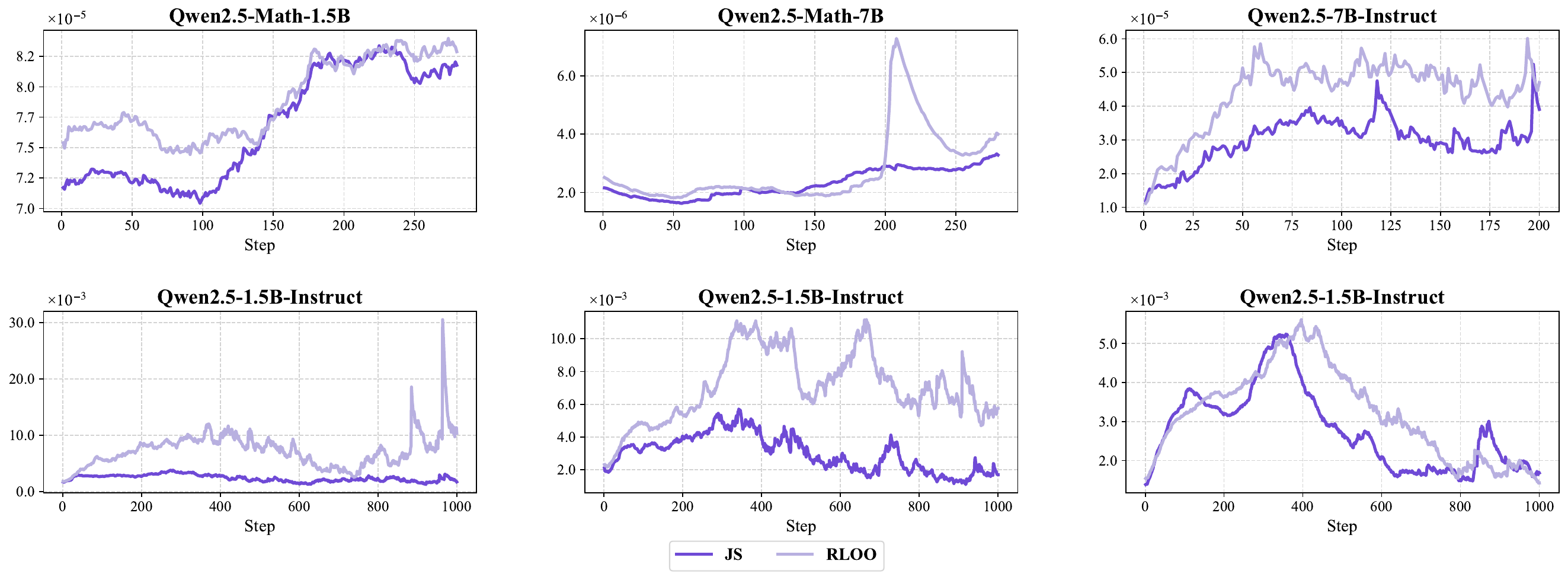}
  \caption{Additional running average gradient variance curves during training.}
  \label{fig:more_variance_curves}
\end{figure}

\clearpage

\subsection{Task Description}
\label{app:task-description}

We describe the tasks used in our experiments in this section.

\subsubsection{Math Reasoning}
\label{app:detail-math}

Our math reasoning dataset consists of 17,917 questions from DAPO17k~\citep{yu2025dapo}. We train models including Qwen2.5-Math-1.5B, Qwen2.5-Math-7B, and Qwen3-4B-Base. An example of the prompt template and its correct response is shown below.

\begin{tcolorbox}[
    colback=gray!5,
    colframe=gray!60,
    coltext=black,
    coltitle=white,
    title={Math Reasoning Example (Qwen2.5-Math-7B)},
    fonttitle=\small\bfseries,
    boxrule=0.4pt,
    arc=2mm,
    left=4pt, right=4pt, top=4pt, bottom=4pt
]

{\ttfamily\small <|im\_start|>system}

{\ttfamily\small Please reason step by step and put the final answer in \textbackslash boxed\{\}. <|im\_end|>}

{\ttfamily\small <|im\_start|>user}

{\ttfamily\small Cities A and B are 45 miles apart. Alicia lives in A and Beth lives in B. Alicia bikes towards B at 18 miles per hour. Leaving at the same time, Beth bikes toward A at 12 miles per hour. How many miles from City A will they be when they meet? Let's think step by step and output the final answer within \textbackslash boxed\{\}. <|im\_end|>}

{\ttfamily\small <|im\_start|>assistant}

{\ttfamily\small To determine where Alicia and Beth will meet, we need to find out how long it will take for them to meet and then use that time to calculate the distance Alicia has traveled.}

{\ttfamily\small 1. First, let's calculate their combined speed. Since Alicia bikes at 18 miles per hour and Beth bikes at 12 miles per hour, their combined speed is 18 + 12 = 30 miles per hour.}

{\ttfamily\small 2. Since the distance between the two cities is 45 miles, the time it takes for them to meet is the total distance divided by their combined speed. So, the time t it takes for them to meet is t = 45/30 = 1.5 hours.}

{\ttfamily\small 3. Now, we need to find out how far Alicia has traveled in 1.5 hours. Since Alicia bikes at 18 miles per hour, the distance she travels in 1.5 hours is 18 * 1.5 = 27 miles.}

{\ttfamily\small So, Alicia and Beth will meet 27 miles from City A. The final answer is \textbackslash boxed\{27\}}

\end{tcolorbox}

\subsubsection{Logic Puzzle Reasoning}

\label{app:detail-logic}
We experiment on three representative logic puzzle tasks: Knights-and-Knaves, Countdown, and Maze. 

\textbf{Knights-and-Knaves (KnK)} is a classic logic puzzle where the goal is to determine truth-tellers (knights) and liars (knaves) based on their statements. We train on two datasets: KnK-Easy, which consists of 10k problems each with 4, 5, or 6 people, and KnK-Hard, which consists of 10k problems each with 6, 7, or 8 people. For each experiment, we evaluate on 300 questions sampled from the same types of KnK puzzles used in training.

\textbf{Countdown} is a numerical game where players use arithmetic operations on given numbers to reach a target value. We train on 10k questions with three numbers and 10k questions with four numbers, and evaluate on 200 questions from the same distribution.

\textbf{Maze} is a spatial puzzle that requires navigating through a complex grid of walls and paths to reach the finish point. We train on 20k 6×6 puzzles where the correct path length ranges from 4 to 8, and evaluate on 200 puzzles from the same distribution.

We provide examples of question content and their corresponding correct responses for each puzzle below.

\begin{tcolorbox}[
    colback=gray!5,
    colframe=gray!60,
    coltext=black,
    coltitle=white,
    title={Knights-and-Knaves},
    fonttitle=\small\bfseries,
    boxrule=0.4pt,
    arc=2mm,
    left=4pt, right=4pt, top=4pt, bottom=4pt
]

{\small\bfseries Question:}

{\ttfamily\small A very special island is inhabited only by sages and fools. Sages always tell the truth, and fools always lie. You meet 6 inhabitants: Owen, Liam, Evelyn, William, Sofia, and Emily. As Owen put it, "William is a sage". Liam was heard saying, "William is a fool". Evelyn noted, "Sofia is a fool and Evelyn is a sage". "if Owen is a sage then Sofia is a sage," William mentioned. "if Liam is a sage then Evelyn is a sage," Sofia claimed. Emily told you that if Emily is a sage then Liam is a fool. So who is a sage and who is a fool? Let's think step by step and output the final answer within \textbackslash boxed\{\}.}

{\small\bfseries Response:}

{\ttfamily\small We need to consider two cases for Owen: (1) If Owen is a sage, then his statement is true, so William must be a sage. (2) If Owen is a fool, then his statement is false, so William must be a fool.}

{\ttfamily\small Liam's statement: "William is a fool." If Liam is a sage, William is a fool. If Liam is a fool, William is a sage.}

{\ttfamily\small William's statement: "If Owen is a sage then Sofia is a sage." Since Sofia is a sage, William's statement is true. Therefore, William must be a sage.}

{\ttfamily\small Owen's statement revisited: "William is a sage." Since William is a sage, Owen's statement is true. Therefore, Owen must be a sage.}

{\ttfamily\small Liam's statement revisited: Since William is a sage, Liam's statement is false. Therefore, Liam must be a fool.}

{\ttfamily\small Emily's statement: "If Emily is a sage then Liam is a fool." Since Emily is a sage, Liam must be a fool. This confirms our previous determination.}

{\ttfamily\small Therefore: Owen is a sage, Liam is a fool, Evelyn is a fool, William is a sage, Sofia is a sage, and Emily is a sage.}

{\ttfamily\small \textbackslash boxed\{Owen is a sage, Liam is a fool, Evelyn is a fool, William is a sage, Sofia is a sage, and Emily is a sage\}}

\end{tcolorbox}

\begin{tcolorbox}[
    colback=gray!5,
    colframe=gray!60,
    coltext=black,
    coltitle=white,
    title={Countdown},
    fonttitle=\small\bfseries,
    boxrule=0.4pt,
    arc=2mm,
    left=4pt, right=4pt, top=4pt, bottom=4pt
]

{\small\bfseries Question:}

{\ttfamily\small Using the numbers [55, 53, 59, 22], create an equation that equals 71. You can use basic arithmetic operations (+, -, *, /) and each number can only be used once. Show your work in <think> </think> tags. And return the final answer in <answer> </answer> tags, for example <answer> (1 + 2) / 3 </answer>.}

{\small\bfseries Response:}

{\ttfamily\small <think>}

{\ttfamily\small We have the numbers 55, 53, 59, and 22. We need an equation that uses each at most once to make 71. Try combinations:}

{\ttfamily\small 55-53-59+22=-45 (not it)}

{\ttfamily\small 55-53+59-22=39 (not it)}

{\ttfamily\small 55+53-59-22=27 (not it)}

{\ttfamily\small 59+22-55-53=-27 (not it)}

{\ttfamily\small 59-22+53-55=35 (not it)}

{\ttfamily\small 53+22+55-59=71 (works!)}

{\ttfamily\small Therefore, an expression that equals 71 is 53+22+55-59.}

{\ttfamily\small </think>}

{\ttfamily\small <answer> 53+22+55-59 </answer>}

\end{tcolorbox}

\begin{tcolorbox}[
    colback=gray!5,
    colframe=gray!60,
    coltext=black,
    coltitle=white,
    title={Maze},
    fonttitle=\small\bfseries,
    boxrule=0.4pt,
    arc=2mm,
    left=4pt, right=4pt, top=4pt, bottom=4pt
]

{\small\bfseries Question:}

{\ttfamily\small You need to solve the following maze. * denotes the wall that you cannot walk through, . denotes available area that you can walk through. S denotes the starting point, E denotes the destination. There are four movement actions: Left (L), Right (R), Up (U), Down (D). Write the final path in <answer> </answer>.}

{\ttfamily\small
\#\# Maze\\
******\\
*.*.E*\\
*....*\\
**.*.*\\
**.S.*\\
******
}

{\small\bfseries Response:}

{\ttfamily\small 1. Locate the start and end: Start S at (5,4), End E at (2,5).}

{\ttfamily\small 2. Check immediate moves from S(5,4): U blocked, D blocked, L open, R open.}

{\ttfamily\small 3. Choose a route to approach E:}

{\ttfamily\small \quad Move R: (5,4)->(5,5), open.}

{\ttfamily\small \quad Move U: (5,5)->(4,5), open.}

{\ttfamily\small \quad Move U: (4,5)->(3,5), open.}

{\ttfamily\small \quad Move U: (3,5)->(2,5), destination reached.}

{\ttfamily\small 4. Validate: Each step lands on open cell until reaching E. No wall crossed.}

{\ttfamily\small <answer> RUUU </answer>}

\end{tcolorbox}

\subsubsection{Multimodal Reasoning}
\label{app:multimodal}

We evaluate on Geometry3k~\citep{lu2021intergps}, a geometry problem dataset that requires visual understanding and mathematical reasoning. Models must interpret figures and solve problems involving shapes, angles, and spatial relationships. Below is an example from the dataset.

\begin{tcolorbox}[
    colback=gray!5,
    colframe=gray!60,
    coltext=black,
    coltitle=white,
    title={Geometry3k},
    fonttitle=\small\bfseries,
    boxrule=0.4pt,
    arc=2mm,
    left=4pt, right=4pt, top=4pt, bottom=4pt
]

{\small\bfseries Question:}

\begin{center}
\includegraphics[width=0.3\textwidth]{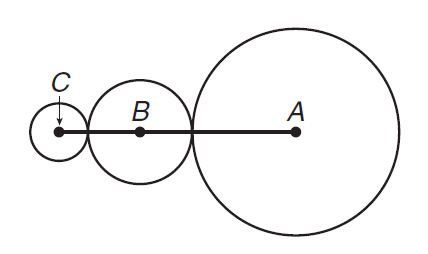}
\end{center}

{\ttfamily\small In the figure, the radius of circle $A$ is twice the radius of circle $B$ and four times the radius of circle $C$. If the sum of the circumferences of the three circles is $42\pi$, find the measure of $\overline{AC}$.}

{\small\bfseries Response:}

{\ttfamily\small Let the radius of circle $C$ be $r$. Then the radius of circle $B$ is $2r$ and the radius of circle $A$ is $4r$.}

{\ttfamily\small The circumference of a circle is $2\pi$ times its radius. So the sum of the circumferences is: $2\pi r + 2\pi(2r) + 2\pi(4r) = 14\pi r$.}

{\ttfamily\small Given that the sum equals $42\pi$: $14\pi r = 42\pi$, so $r = 3$.}

{\ttfamily\small Since the circles are externally tangent, $\overline{AC}$ equals the sum of the radii of all three circles: $\overline{AC} = r + 2r + 2r + 4r = 9r = 27$.}

{\ttfamily\small The answer is \textbackslash boxed\{27\}.}

\end{tcolorbox}

\clearpage

\subsection{Experimental Setups}
\label{app:exp-setups}

We summarize the training configurations for all experiments in this section.

\begin{table}[h]
  \centering
  \label{tab:config-1.5b}
  \begin{tcolorbox}[
    enhanced, hbox,
    title={Training setup of Qwen2.5-Math-1.5B},
    colback=teal!5, colframe=teal, coltext=black, coltitle=white,
    fonttitle=\bfseries, arc=1mm, boxrule=1pt, boxsep=1pt,
    left=2pt, right=2pt, top=0pt, bottom=2pt,
    toptitle=3pt, bottomtitle=3pt, center
  ]
    \small
    \renewcommand{\arraystretch}{1.2}
    \setlength{\tabcolsep}{8pt}
    \begin{tabular}{l|l|l|l}
      \textbf{Parameter} & \textbf{Value} & \textbf{Parameter} & \textbf{Value} \\
      \hline
      Base model & Qwen2.5-Math-1.5B & Training set & DAPO17k \\
      Test set & MATH500, AMC23, Olympiad & Prompts per batch & 64 \\
      Generations per prompt & 4 & Grad updates per step & 2 \\
      Max prompt len & 1024 & Max response len & 2048 \\
      Learning rate & $2 \times 10^{-6}$ & Clip ratio & 0.22 \\
      KL coeff & 0.0 & Entropy coeff & 0.0 \\
      Rollout temp & 0.8 & Validation temp & 0.8 \\
      Validation samples & 16 & Validation interval & 50 steps \\
      Lora Rank & 0 & Device & 4$\times$GH200 \\
    \end{tabular}
  \end{tcolorbox}
\end{table}

\begin{table}[h]
  \centering
  \label{tab:config-7b}
  \begin{tcolorbox}[
    enhanced, hbox,
    title={Training setup of Qwen2.5-Math-7B},
    colback=teal!5, colframe=teal, coltext=black, coltitle=white,
    fonttitle=\bfseries, arc=1mm, boxrule=1pt, boxsep=1pt,
    left=2pt, right=2pt, top=0pt, bottom=2pt,
    toptitle=3pt, bottomtitle=3pt, center
  ]
    \small
    \renewcommand{\arraystretch}{1.2}
    \setlength{\tabcolsep}{8pt}
    \begin{tabular}{l|l|l|l}
      \textbf{Parameter} & \textbf{Value} & \textbf{Parameter} & \textbf{Value} \\
      \hline
      Base model & Qwen2.5-Math-7B & Training set & DAPO17k \\
      Test set & MATH500, AMC23, Olympiad & Prompts per batch & 64 \\
      Generations per prompt & 4 & Grad updates per step & 2 \\
      Max prompt len & 1024 & Max response len & 2048 \\
      Learning rate & $2 \times 10^{-5}$ & Clip ratio & 0.22 \\
      KL coeff & 0.0 & Entropy coeff & 0.0 \\
      Rollout temp & 0.8 & Validation temp & 0.8 \\
      Validation samples & 16 & Validation interval & 50 steps \\
      Lora Rank & 256 & Device & 4$\times$GH200 \\
    \end{tabular}
  \end{tcolorbox}
\end{table}

\begin{table}[h]
  \centering
  \label{tab:config-4b}
  \begin{tcolorbox}[
    enhanced, hbox,
    title={Training setup of Qwen3-4B-Base},
    colback=teal!5, colframe=teal, coltext=black, coltitle=white,
    fonttitle=\bfseries, arc=1mm, boxrule=1pt, boxsep=1pt,
    left=2pt, right=2pt, top=0pt, bottom=2pt,
    toptitle=3pt, bottomtitle=3pt, center
  ]
    \small
    \renewcommand{\arraystretch}{1.2}
    \setlength{\tabcolsep}{8pt}
    \begin{tabular}{l|l|l|l}
      \textbf{Parameter} & \textbf{Value} & \textbf{Parameter} & \textbf{Value} \\
      \hline
      Base model & Qwen3-4B-Base & Training set & DAPO17k \\
      Test set & MATH500, AMC23, Olympiad & Prompts per batch & 64 \\
      Generations per prompt & 4 & Grad updates per step & 2 \\
      Max prompt len & 1024 & Max response len & 3072 \\
      Learning rate & $2 \times 10^{-6}$ & Clip ratio & 0.28 \\
      KL coeff & 0.0 & Entropy coeff & 0.0 \\
      Rollout temp & 1.0 & Validation temp & 1.0 \\
      Validation samples & 16 & Validation interval & 50 steps \\
      Lora Rank & 0 & Device & 4$\times$H100 \\
    \end{tabular}
  \end{tcolorbox}
\end{table}

\begin{table}[h]
  \centering
  \label{tab:config-knk-easy}
  \begin{tcolorbox}[
    enhanced, hbox,
    title={Training setup for KnK-Easy},
    colback=teal!5, colframe=teal, coltext=black, coltitle=white,
    fonttitle=\bfseries, arc=1mm, boxrule=1pt, boxsep=1pt,
    left=2pt, right=2pt, top=0pt, bottom=2pt,
    toptitle=3pt, bottomtitle=3pt, center
  ]
    \small
    \renewcommand{\arraystretch}{1.2}
    \setlength{\tabcolsep}{8pt}
    \begin{tabular}{l|l|l|l}
      \textbf{Parameter} & \textbf{Value} & \textbf{Parameter} & \textbf{Value} \\
      \hline
      Base model & Qwen2.5-7B-Instruct & Training set & KnK-4, KnK-5, KnK-6 \\
      Test set & KnK-4/5/6-Test & Prompts per batch & 32 \\
      Generations per prompt & 8 & Grad updates per step & 2 \\
      Max prompt len & 1024 & Max response len & 2048 \\
      Learning rate & $4 \times 10^{-5}$ & Clip ratio & 0.22 \\
      KL coeff & 0.0 & Entropy coeff & 0.0 \\
      Rollout temp & 0.7 & Validation temp & 0.7 \\
      Validation samples & 16 & Validation interval & 25 steps \\
      Lora Rank & 256 & Device & 4$\times$GH200 \\
    \end{tabular}
  \end{tcolorbox}
\end{table}

\begin{table}[h]
  \centering
  \label{tab:config-knk-hard}
  \begin{tcolorbox}[
    enhanced, hbox,
    title={Training setup for KnK-Hard},
    colback=teal!5, colframe=teal, coltext=black, coltitle=white,
    fonttitle=\bfseries, arc=1mm, boxrule=1pt, boxsep=1pt,
    left=2pt, right=2pt, top=0pt, bottom=2pt,
    toptitle=3pt, bottomtitle=3pt, center
  ]
    \small
    \renewcommand{\arraystretch}{1.2}
    \setlength{\tabcolsep}{8pt}
    \begin{tabular}{l|l|l|l}
      \textbf{Parameter} & \textbf{Value} & \textbf{Parameter} & \textbf{Value} \\
      \hline
      Base model & Qwen2.5-7B-Instruct & Training set & KnK-6, KnK-7, KnK-8 \\
      Test set & KnK-6/7/8-Test & Prompts per batch & 32 \\
      Generations per prompt & 8 & Grad updates per step & 2 \\
      Max prompt len & 1024 & Max response len & 2048 \\
      Learning rate & $4 \times 10^{-5}$ & Clip ratio & 0.22 \\
      KL coeff & 0.0 & Entropy coeff & 0.0 \\
      Rollout temp & 0.7 & Validation temp & 0.7 \\
      Validation samples & 16 & Validation interval & 25 steps \\
      Lora Rank & 256 & Device & 4$\times$GH200 \\
    \end{tabular}
  \end{tcolorbox}
\end{table}

\begin{table}[h]
  \centering
  \label{tab:config-countdown}
  \begin{tcolorbox}[
    enhanced, hbox,
    title={Training setup for Countdown},
    colback=teal!5, colframe=teal, coltext=black, coltitle=white,
    fonttitle=\bfseries, arc=1mm, boxrule=1pt, boxsep=1pt,
    left=2pt, right=2pt, top=0pt, bottom=2pt,
    toptitle=3pt, bottomtitle=3pt, center
  ]
    \small
    \renewcommand{\arraystretch}{1.2}
    \setlength{\tabcolsep}{8pt}
    \begin{tabular}{l|l|l|l}
      \textbf{Parameter} & \textbf{Value} & \textbf{Parameter} & \textbf{Value} \\
      \hline
      Base model & Qwen2.5-3B & Training set & Countdown3, Countdown4 \\
      Test set & Countdown3/4-Test & Prompts per batch & 64 \\
      Generations per prompt & 5 & Grad updates per step & 2 \\
      Max prompt len & 512 & Max response len & 1024 \\
      Learning rate & $1 \times 10^{-6}$ & Clip ratio & 0.22 \\
      KL coeff & 0.0 & Entropy coeff & 0.0 \\
      Rollout temp & 0.7 & Validation temp & 0.7 \\
      Validation samples & 16 & Validation interval & 10 steps \\
      Lora Rank & 0 & Device & 2$\times$GH200 \\
    \end{tabular}
  \end{tcolorbox}
\end{table}

\begin{table}[h]
  \centering
  \label{tab:config-maze}
  \begin{tcolorbox}[
    enhanced, hbox,
    title={Training setup for Maze},
    colback=teal!5, colframe=teal, coltext=black, coltitle=white,
    fonttitle=\bfseries, arc=1mm, boxrule=1pt, boxsep=1pt,
    left=2pt, right=2pt, top=0pt, bottom=2pt,
    toptitle=3pt, bottomtitle=3pt, center
  ]
    \small
    \renewcommand{\arraystretch}{1.2}
    \setlength{\tabcolsep}{8pt}
    \begin{tabular}{l|l|l|l}
      \textbf{Parameter} & \textbf{Value} & \textbf{Parameter} & \textbf{Value} \\
      \hline
      Base model & Ministral-8B-Instruct & Training set & Maze6x6 \\
      Test set & Maze6x6-Test & Prompts per batch & 32 \\
      Generations per prompt & 8 & Grad updates per step & 2 \\
      Max prompt len & 1024 & Max response len & 2048 \\
      Learning rate & $3 \times 10^{-7}$ & Clip ratio & 0.25 \\
      KL coeff & 0.0 & Entropy coeff & 0.0 \\
      Rollout temp & 0.7 & Validation temp & 0.7 \\
      Validation samples & 16 & Validation interval & 25 steps \\
      Lora Rank & 0 & Device & 4$\times$GH200 \\
    \end{tabular}
  \end{tcolorbox}
\end{table}

\begin{table}[h]
  \centering
  \label{tab:exp-config}
  \begin{tcolorbox}[
    enhanced, hbox,
    title={Training Setup for GSM8k},
    colback=teal!5, colframe=teal, coltext=black, coltitle=white,
    fonttitle=\bfseries, arc=1mm, boxrule=1pt, boxsep=1pt,
    left=2pt, right=2pt, top=0pt, bottom=2pt,
    toptitle=3pt, bottomtitle=3pt, center
  ]
    \small
    \renewcommand{\arraystretch}{1.2}
    \setlength{\tabcolsep}{8pt}
    \begin{tabular}{l|l|l|l}
      \textbf{Parameter} & \textbf{Value} & \textbf{Parameter} & \textbf{Value} \\
      \hline
      Base model & Qwen2.5-0.5B-Instruct & Training set & GSM8k-Train \\
      Test set & GSM8k-Test & Prompts per batch & 64 \\
      Generations per prompt & 4 & Grad updates per step & 1 \\
      Max prompt len & 1024 & Max response len & 2048 \\
      Learning rate & $1 \times 10^{-6}$ & Clip ratio & 0.2 \\
      KL coeff & 0.0 & Entropy coeff & 0.0 \\
      Rollout temp & 0.7 & Validation temp & 0.5 \\
      Validation samples & 10 & Validation interval & 100 steps \\
      & & Device & 1$\times$GH200 \\
    \end{tabular}
  \end{tcolorbox}
\end{table}

\begin{table}[h]
  \centering
  \label{tab:config-3b-vl}
  \begin{tcolorbox}[
    enhanced, hbox,
    title={Training setup of Qwen2.5-VL-3B-Instruct},
    colback=teal!5, colframe=teal, coltext=black, coltitle=white,
    fonttitle=\bfseries, arc=1mm, boxrule=1pt, boxsep=1pt,
    left=2pt, right=2pt, top=0pt, bottom=2pt,
    toptitle=3pt, bottomtitle=3pt, center
  ]
    \small
    \renewcommand{\arraystretch}{1.2}
    \setlength{\tabcolsep}{8pt}
    \begin{tabular}{l|l|l|l}
      \textbf{Parameter} & \textbf{Value} & \textbf{Parameter} & \textbf{Value} \\
      \hline
      Base model & Qwen2.5-VL-3B-Instruct & Training set & Geo3k \\
      Test set & Geo3k-Test & Prompts per batch & 512 \\
      Generations per prompt & 5 & Grad updates per step & 4 \\
      Max prompt len & 1024 & Max response len & 1024 \\
      Learning rate & $1 \times 10^{-6}$ & Clip ratio & 0.2 \\
      KL coeff & 0.01 & Entropy coeff & 0.0 \\
      Rollout temp & 0.7 & Validation temp & 0.7 \\
      Validation samples & 16 & Validation interval & 20 steps \\
      Lora Rank & 0 & Device & 4$\times$GH200 \\
    \end{tabular}
  \end{tcolorbox}
\end{table}

\begin{table}[h]
  \centering
  \label{tab:config-7b-vl}
  \begin{tcolorbox}[
    enhanced, hbox,
    title={Training setup of Qwen2.5-VL-7B-Instruct},
    colback=teal!5, colframe=teal, coltext=black, coltitle=white,
    fonttitle=\bfseries, arc=1mm, boxrule=1pt, boxsep=1pt,
    left=2pt, right=2pt, top=0pt, bottom=2pt,
    toptitle=3pt, bottomtitle=3pt, center
  ]
    \small
    \renewcommand{\arraystretch}{1.2}
    \setlength{\tabcolsep}{8pt}
    \begin{tabular}{l|l|l|l}
      \textbf{Parameter} & \textbf{Value} & \textbf{Parameter} & \textbf{Value} \\
      \hline
      Base model & Qwen2.5-VL-7B-Instruct & Training set & Geo3k \\
      Test set & Geo3k-Test & Prompts per batch & 512 \\
      Generations per prompt & 5 & Grad updates per step & 4 \\
      Max prompt len & 1024 & Max response len & 1024 \\
      Learning rate & $1 \times 10^{-6}$ & Clip ratio & 0.2 \\
      KL coeff & 0.01 & Entropy coeff & 0.0 \\
      Rollout temp & 0.7 & Validation temp & 0.7 \\
      Validation samples & 16 & Validation interval & 20 steps \\
      Lora Rank & 0 & Device & 4$\times$GH200 \\
    \end{tabular}
  \end{tcolorbox}
\end{table}

\begin{table}[h]
  \centering
  \label{tab:config-1.5b-it}
  \begin{tcolorbox}[
    enhanced, hbox,
    title={Training setup of Qwen2.5-1.5B-Instruct},
    colback=teal!5, colframe=teal, coltext=black, coltitle=white,
    fonttitle=\bfseries, arc=1mm, boxrule=1pt, boxsep=1pt,
    left=2pt, right=2pt, top=0pt, bottom=2pt,
    toptitle=3pt, bottomtitle=3pt, center
  ]
    \small
    \renewcommand{\arraystretch}{1.2}
    \setlength{\tabcolsep}{8pt}
    \begin{tabular}{l|l|l|l}
      \textbf{Parameter} & \textbf{Value} & \textbf{Parameter} & \textbf{Value} \\
      \hline
      Base model & Qwen2.5-1.5B-Instruct & Training set & KnK-2, KnK-3 \\
      Test set & KnK-2/3-Test & Prompts per batch & 32 \\
      Generations per prompt & 8 & Grad updates per step & 2 \\
      Max prompt len & 1024 & Max response len & 2048 \\
      Learning rate & $5 \times 10^{-7}$ & Clip ratio & 0.22 \\
      KL coeff & 0.0 & Entropy coeff & 0.0 \\
      Rollout temp & 0.7 & Validation temp & 0.7 \\
      Validation samples & 16 & Validation interval & 25 steps \\
      Lora Rank & 0 & Device & 4$\times$H100 \\
    \end{tabular}
  \end{tcolorbox}
\end{table}

\clearpage


\end{document}